%% file: main.tex
\documentclass[sigconf, 9pt]{acmart}

\usepackage{caption}
\usepackage{graphicx}
\usepackage{xspace}
\usepackage{cleveref}
\usepackage{lipsum}
\usepackage[normalem]{ulem}
\usepackage{enumitem}
\usepackage{geometry}
\usepackage{amsmath}
\setlist[itemize]{noitemsep, topsep=0pt}
\usepackage[utf8]{inputenc}
\usepackage[T1]{fontenc}
\usepackage{adjustbox}
\usepackage{subfigure}
\usepackage{multirow}
\usepackage[normalem]{ulem}

\usepackage{soul}
\usepackage{hyperref}
\hypersetup{
    colorlinks=true,
    linkcolor=blue,
    filecolor=magenta,
    urlcolor=cyan,
}

\usepackage{siunitx}
\usepackage{tikz}
\newcommand*\circled[1]{\tikz[baseline=(char.base)]{
            \node[shape=circle,draw,inner sep=0.8pt] (char) {#1};}}

\makeatletter
\DeclareRobustCommand\onedot{\futurelet\@let@token\@onedot}
\def\@onedot{\ifx\@let@token.\else.\null\fi\xspace}

\makeatother

\AtBeginDocument{%
  \providecommand\BibTeX{{%
    \normalfont B\kern-0.5em{\scshape i\kern-0.25em b}\kern-0.8em\TeX}}}

\newcommand{\rev}[1]{\textcolor{black}{#1}}
\newcommand{\urlcolor}[1]{\textcolor{magenta}{#1}}
\newcommand{\system}{\textit{HypeMeFed}\xspace}

\copyrightyear{2024}
\acmYear{2024}
\setcopyright{acmlicensed}
\acmConference[Sensys '24]{The 22nd ACM Conference on Embedded Networked Sensor Systems}{November 4--7, 2024, Hangzhou, China}{Hangzhou, China}
\acmBooktitle{The 22nd ACM Conference on Embedded Networked Sensor Systems (SenSys' 24), November 4--7, 2024, Hangzhou, China}
\acmPrice{}
\acmDOI{}
\acmISBN{}

\begin{document}
\title[]{Effective Heterogeneous Federated Learning \textit{via} Efficient Hypernetwork-based Weight Generation}


\author{Yujin Shin*}
\thanks{*Both authors contributed equally to this research.}
\email{yujin_shin@yonsei.ac.kr}
\affiliation{%
  \institution{Yonsei University}
  \country{}
}
\author{Kichang Lee*}
\email{kichang.lee@yonsei.ac.kr}
\affiliation{%
  \institution{Yonsei University}
  \country{}
}
\author{Sungmin Lee}
\email{i.am.sungmin.lee@yonsei.ac.kr}
\affiliation{%
  \institution{Yonsei University}
  \country{}
}

\author{You Rim Choi}
\email{yrchoi@snu.ac.kr}
\affiliation{%
  \institution{Seoul National University}
  \country{}
}

\author{Hyung-Sin Kim}
\email{hyungkim@snu.ac.kr}
\affiliation{%
  \institution{Seoul National University}
  \country{}
}

\author{JeongGil Ko}
\email{jeonggil.ko@yonsei.ac.kr}
\affiliation{%
  \institution{Yonsei University, POSTECH}
  \country{}
}


\begin{abstract}
\input{sec/0_abstract}

\end{abstract}


\settopmatter{printfolios=True} 
\settopmatter{printacmref=True}
\maketitle
\input{sec/1_intro}

\input{sec/2_relwork}

\input{sec/3_prelim}

\input{sec/4_design}

\input{sec/5_eval}

\input{sec/6_discussion}

\input{sec/7_conclusion}

\begin{acks}
We thank the anonymous reviewers and shepherd for their efforts in providing constructive comments for the paper. The authors would also like to thank Jaeho Jin for help in configuring the testbed environment and Sinyoung Oh for designing the figures. This work was supported by the Korean Ministry of Science and ICT (MSIT) and IITP under grants IITP-2024-2020-0-01461 (ITRC Program) and IITP-2022-0-00420. Hyung-sin Kim and JeongGil Ko are the corresponding authors for this work.
\end{acks}


\balance
\bibliographystyle{ACM-Reference-Format}
\bibliography{reference}

\clearpage

\end{document}

%% file: sec/0_abstract.tex
%
While federated learning leverages distributed client resources, it faces challenges due to heterogeneous client capabilities. This necessitates allocating models suited to clients' resources and careful parameter aggregation to accommodate this heterogeneity. We propose \system, a novel federated learning framework for supporting client heterogeneity by combining a multi-exit network architecture with hypernetwork-based model weight generation. This approach aligns the feature spaces of heterogeneous model layers and resolves per-layer information disparity during weight aggregation.
To practically realize \system, we also propose a low-rank factorization approach to minimize computation and memory overhead associated with hypernetworks. 
Our evaluations on a real-world heterogeneous device testbed indicate that \system enhances accuracy by 5.12\% over FedAvg, reduces the hypernetwork memory requirements by 98.22\%, and accelerates its operations by 1.86$\times$ compared to a naive hypernetwork approach. These results demonstrate \system's effectiveness in leveraging and engaging heterogeneous clients for federated learning.

%% file: sec/1_intro.tex
\section{Introduction}
\label{sec:intro}

The proliferation of deep neural networks fueled by extensive datasets has significantly enhanced mobile and IoT applications, ranging from remote physiological signal monitoring~\cite{park2020heartquake, shao2024joey}, human action recognition~\cite{xu2023mesen, jiang2018towards}, and audio recognition~\cite{coucke2018snips, yang2023pdassess}. Smart devices such as smartphones, smartwatches, and wearable sensors capture high-fidelity data, enabling robust machine learning model development~\cite{stisen2015smart, pham2022pros}. These advancements support diverse real-world applications~\cite{park2021enabling, yun2024powdew, huynh2019vitamon}. However, client data often contains sensitive information, and the high communication costs of transmitting raw data to a central server pose challenges for creating comprehensive large-scale datasets suitable for centralized deep model training.


Recently, federated learning has emerged as a promising approach to building effective machine learning models without explicitly sharing private data. Instead, it shares locally trained model parameters, indirectly transferring local knowledge in a distributed and collaborative manner~\cite{mcmahan2017communication}. Naturally, federated learning has catalyzed advancements in on-device training~\cite{gim2022memory, lin2022device, wang2022melon}, enabling a diverse array of real-world applications, including healthcare, smart cities, and autonomous driving~\cite{silva2020fed, ouyang2023harmony}.

Despite extensive research in federated learning, most prior works assume a homogeneous model architecture to be shared across all clients~\cite{mcmahan2017communication, li2021fedmask}. This assumption overlooks the diverse range of devices participating in federated learning (i.e., device heterogeneity), especially in practical cross-platform settings. Devices can vary significantly in processing power, memory capacity, and available energy~\cite{park2024fedhm}. This issue of computational resource heterogeneity is even more pronounced in mobile and embedded applications, where clients use devices with vastly different computational capabilities, from smartphones to embedded IoT platforms~\cite{cho2022flame}.

Without considering client-side computational resource heterogeneity, model design is constrained to the capabilities of the least powerful client, leading to suboptimal performance. Furthermore, clients with weaker computing capabilities may struggle to meet training deadlines, raising fairness concerns when excluded due to timeouts~\cite{shin2022fedbalancer, li2020federated} or can delay the overall federated learning process as the server waits for clients lagging behind~\cite{ouyang2023harmony, park2024fedhm}. Therefore, addressing device heterogeneity in federated learning is crucial to balance effective model building and achieve efficient training times~\cite{li2020lotteryfl, diao2020heterofl}. To maximize the distributed computing resources, a federated learning framework must adapt its model capacity to allow all clients to participate in the federated learning process with a suitable model~\cite{alam2022fedrolex, ilhan2023scalefl}.

\rev{Existing solutions for device heterogeneity in federated learning include techniques like knowledge distillation~\cite{li2019fedmd, he2020group}, local model pruning~\cite{li2021hermes, li2020lotteryfl}, and adaptive model scaling~\cite{ilhan2023scalefl, diao2020heterofl}. Knowledge distillation requires a well-balanced public dataset for transferring knowledge from larger models, which is often impractical in privacy-sensitive settings. Local model pruning reduces resource demands, but still requires full-model training in the initial phase, which is infeasible for many resource-limited devices. Adaptive model scaling, either width-wise or depth-wise, adjusts model sizes based on client resource availablity, but faces the issue of information imbalance during server aggregation, where smaller models provide less informative updates~\cite{shen2024fedconv, yao2021fedhm}. These limitations highlight the need for approaches that balance resource optimization with effective neural network parameter aggregation in heterogeneous environments.}
This work builds on adaptive model scaling, where neural networks can be split into varying depths and clients can utilize differently sized subnetworks of the original model based on their computational capabilities. To achieve this, we adopt a \textit{multi-exit model architecture}, integrating intermediate exits to split a larger model depth-wise. This ensures that heterogeneous clients can align their feature spaces across layers, facilitating consistent context sharing during parameter aggregation (Sec.~\ref{subsec:multiexitdesign}).


With heterogeneous platforms utilizing multi-exit architectures, all clients execute the model up to the first exit, while only a subset of powerful clients proceed to the final. This results in varying sample availability (i.e., information disparity) across different exits at the server, particularly affecting deeper layers with fewer samples for aggregation. Our approach mitigates this challenge using \textit{hypernetworks} to infer and generate weights for layers lacking sufficient samples, enhancing model aggregation effectiveness. However, despite hypernetworks operating on the server, their resource requirements are extremely high; resulting in large memory usage and long computation times. \rev{We emphasize that hypernetworks can struggle to run effectively even on powerful server platforms due to their memory requirements~\cite{ivison2022hint, chauhan2024brief}.} To address this, we propose a \textit{Low-Rank Factorization (LRF)-based compression scheme for neural network parameters}, aimed at optimizing hypernetwork operations for federated learning (Sec.~\ref{subsec:hypernetdesign}). \rev{This LRF-based approach ensures that hypernetworks can generate parameters efficiently at the server, making \system scalable and deployable in real-world heterogeneous settings.}

We integrate these techniques as a heterogeneous federated learning framework, \system, and present extensive evaluations via GPU-based emulations and a real-world testbed of heterogeneous embedded platforms. \system outperforms all state-of-the-art baselines, and our testbed results show that it improves model accuracy by \textbf{5.13\%} compared to FedAvg. Furthermore, with its LRF-based optimization, \system reduces the hypernetwork's memory requirement by \textbf{98.22\%} and achieves \textbf{1.86$\times$} faster hypernetwork operations compared to a naive approach. Overall, our results suggest that \system is an effective solution to enable federated learning for heterogeneous clients. We provide an open-sourced implementation of \system at \urlcolor{\url{https://github.com/eis-lab/HypeMeFed}}.

The contributions of this work are three-fold as follows:

\begin{itemize}[leftmargin=*]
    \item We identify practical challenges in federated learning for heterogeneous clients and propose \rev{\system, a novel framework that combines a multi-exit neural network architecture with hypernetwork-based model weight generation. This approach supports effective federated learning across clients with diverse computational capabilities by addressing the issue of layer-specific information disparity.}

    \item While hypernetworks offer an environment to generate model weights for portions of the network that lack client-sourced parameters, their naive use results in significant computational and memory overhead. \rev{We introduce an LRF-based compression scheme to optimize hypernetwork operations, significantly reducing resource demands and enabling practical deployment of hypernetworks in federated learning.}

    \item We perform extensive evaluations using both GPU-based emulations and embedded platform implementations to show the effectiveness of \system. Our results suggest that \system is a practical solution for enabling federated learning over clients with heterogeneous capabilities.
    
\end{itemize}


%% file: sec/2_relwork.tex
\section{Related work and Background}
\label{sec:relwork}

We first positioning our work within current literature and discuss foundational concepts and methodologies relevant to our study. 

\vspace{0.1in}
\noindent\textbf{Device Heterogeneity in Federated Learning.} While federated learning research has mostly focused on homogeneous models with little consideration of client-side computing capabilities~\cite{diao2020heterofl}, supporting federated learning for heterogeneous clients is crucial. It allows for tailored models that adapt to diverse computational capabilities, data distributions, and privacy requirements of various devices~\cite{zhang2022federated, bonawitz2019towards, li2020federated}. Heterogeneous federated learning leverages each client's unique capabilities, ensuring that even resource-constrained devices contribute effectively to enhance model robustness, maximize the potential of collective data, and lead to better generalization across tasks and environments~\cite{lin2020ensemble}.

\rev{One early approach to achieve device heterogeneity in federated learning is federated distillation (FD)~\cite{li2019fedmd, seo202216}, which leverages knowledge distillation~\cite{hinton2015distilling}. Instead of sharing model weights, FD aggregates class scores from a large public dataset computed locally at each client to accommodate diverse client models. However, FD requires the server to maintain access to a well-balanced public shared dataset, which can be impractical in privacy-sensitive settings. Furthermore, the availability of such datasets may also introduce biases, limiting model the generalizability across diverse client data distributions.}

An alternative approach involves pruning local models based on the Lottery Ticket hypothesis~\cite{frankle2018lottery}. In LotteryFL~\cite{li2020lotteryfl} and Hermes~\cite{li2021hermes}, each client learns a subnetwork of the global model (i.e., lottery ticket network) through iterative pruning, rewinding, and training, exchanging only sparse pruned parameters. However, this iterative process is practically challenging for resource-limited clients, as they must initially train the heavy full model to determine which weights to prune. This requires a large number of rounds for model convergence and can cause training delays. \rev{Furthermore, aggressively pruned models may lose significant amounts of information, affecting both model convergence and final accuracy.} 

A different approach reduces model capacity by scaling width, depth, or both, aligning with individual client conditions based on available resources. In width-based scaling (e.g., HeteroFL~\cite{diao2020heterofl}, FedRolex~\cite{alam2022fedrolex}), clients subsample different channel parts per layer for server aggregation. HeteroFL applies static subsampling across global rounds, while FedRolex employs a rolling method to distribute channel parameters evenly among clients by rotating the sampling frame. 

In contrast, depth-based scaling methods like DepthFL~\cite{kim2023depthfl}, REEFL~\cite{lee2024recurrent}, and the hybrid method ScaleFL~\cite{ilhan2023scalefl} adjust models in depth using a multi-exit architecture. ScaleFL further improves shallower models with self-distillation from deeper models. 
Despite their superior performance compared to width-based methods, they suffer from information disparity between shallower and deeper layers. With clients participating in aggregation using varying model depths, shallower layers receive more samples, amplifying disparity in information, representation capacity, and contribution to the global model. This information disparity not only harms the generalizability of the global model but also undermines self-distillation, as students (shallower models) learn with more data than teachers (deeper models), which is atypical.


This motivates \system{} to propose an alternative method to mitigate information disparity. Additionally, existing depth-scaling methods lack justification for multi-exit architectures and comprehensive analysis of information disparities. This work provides an in-depth analysis in Section~\ref{subsec:3.1} and designs a federated framework based on our findings.

\vspace{0.1in}

\noindent\textbf{Multi-Exit Network Architecture.} As the name implies, multi-exit architectures, inject multiple (early) exit points within a neural network to enhance inference latency by enabling adaptive inference and eliminating unnecessary operations~\cite{laskaridis2021adaptive, lee2024recurrent}. Teerapittayanon et al. introduced BranchNet, a pioneering multi-exit deep neural network featuring multiple classifiers integrated within the baseline model to improve computational latency~\cite{teerapittayanon2016branchynet}. Given that mobile and embedded devices often experience lagging inference due to resource limitations, multi-exit architectures have been often applied in various real-world sensing applications within mobile and embedded computing environments~\cite{kouris2022adaptable, leontiadis2021s, kim20200}. \rev{However, while these existing methods can improve performance during inference, they hold challenges in local training, as all exits must be trained. As a result, heterogeneous clients cannot benefit from this approach for the training phase.}

\vspace{0.1in}

\noindent\textbf{Hypernetworks.} Hypernetworks are neural networks designed to generate weights for another neural network~\cite{ha2016hypernetworks}, offering flexibility in applications like computer vision~\cite{jia2016dynamic, klocek2019hypernetwork} and 3D scene representation~\cite{littwin2019deep, sitzmann2020implicit}. In federated learning, hypernetworks are used to enhance model initialization and personalization~\cite{shamsian2021personalized, litany2022federated} and to determine aggregation ratios for model personalization~\cite{ma2022layer}. \rev{However, in existing federated learning using hypernetworks, the training of the hypernetwork takes place on each local client, imposing an additional computational burden on the clients.} 

Our work proposes a novel approach by designing and leveraging efficient hypernetworks specifically to address challenges posed by heterogeneous federated learning scenarios. Unlike previous works that exploit hypernetworks for general model initialization or focus on model personalization, our approach aims to dynamically generate missing model parameters to address the information disparity problem in heterogeneous federated learning. Furthermore, we introduce a model compression-based hypernetwork approach to minimize the computational and memory overhead associated with hypernetwork operations.

%% file: sec/3_prelim.tex
\vspace{-1ex}
\section{Challenges and Preliminary Study}
\label{sec:prelim}


The recently evolving paradigm of distributed training via federated learning offers a novel approach to developing effective models by utilizing data collected on end devices such as IoT and mobile platforms. Federated learning method enables the system to fully exploit data from many clients while maintaining a privacy-aware environment by ensuring that raw data remains on the client's devices.

To fully leverage this distributed data, it is crucial to maximize the number of clients participating in the federated learning process. However, studies highlight that opportunities for participation are often limited due to various factors. A major obstacle is the heterogeneity of computing resources, which can limit both model performance and client participation. It is unrealistic to assume that all clients have similar computing resources and can train the same model, as clients with limited resources (e.g., CPU, RAM, GPU) may be unable to join the training process, despite possessing valuable data. Conversely, if the server employs a lightweight model to accommodate as many clients as possible, the performance of clients with ample resources can be constrained by the limited model capacity. Additionally, less capable clients may require extensive training time, potentially exceeding the server's deadline and preventing their participation.

\vspace{-1ex}
\subsection{Challenges and Potential Solutions}
\label{subsec:3.1}

To accommodate the diverse range of computing resources in clients with heterogeneous capabilities, a server can configure and allocate models with different levels of capacity (i.e., number of parameters) suitable for each client. A model with a relatively small number of parameters generally requires less memory, computation, and latency for both inference and training~\cite{horvath2021fjord}. However, doing so in federated learning presents a challenge: the server must aggregate knowledge (i.e., parameters) from locally trained models with non-identical architectures. One feasible approach is to exploit subnetwork architectures of the full neural network by splitting the original network in a depth-wise manner (i.e., layer-level split). While promising, given that heterogeneous models maintain some level of architectural commonness, we identify two notable issues with the naive approach of simply dividing the network in a depth-wise manner.

\noindent \textbf{[Issue 1] Misaligned Feature Space:}
The first issue with depth-wise network splits is the \textit{misaligned feature space}. Layers closer to the input focus on local features, while layers closer to the output capture global features for tasks like classification. When devices with varying capabilities hold different ``splits'' of the network, the feature spaces for the same layer can differ due to their relative positions in subnetworks. For example, a layer might act as an intermediate layer in the full model but as a final layer in a subnetwork. This misalignment can significantly harm model performance in federated learning, as also reported by Luo et al.~\cite{luo2021no}.

\noindent \textbf{[Solution 1] Multi-exit Architecture:}
To better align the representation space in each subnetwork, we leverage a \textit{multi-exit architecture}~\cite{teerapittayanon2016branchynet, laskaridis2021adaptive}. This approach adds intermediate exit points and enables each depth-wise split subnetwork to independently learn global features while also exploiting them as local features for subsequent subnetworks. With a multi-exit architecture, we can improve feature alignment in models with different layer depths. This ensures that each layer contributes to both local and global feature learning, enhancing the overall performance and consistency of the federated learning process. We will discuss more about the multi-exit architecture and its implementation in Section~\ref{sec:design}.

\begin{figure}[!t]
    \centering
    \includegraphics[width=0.8\linewidth]{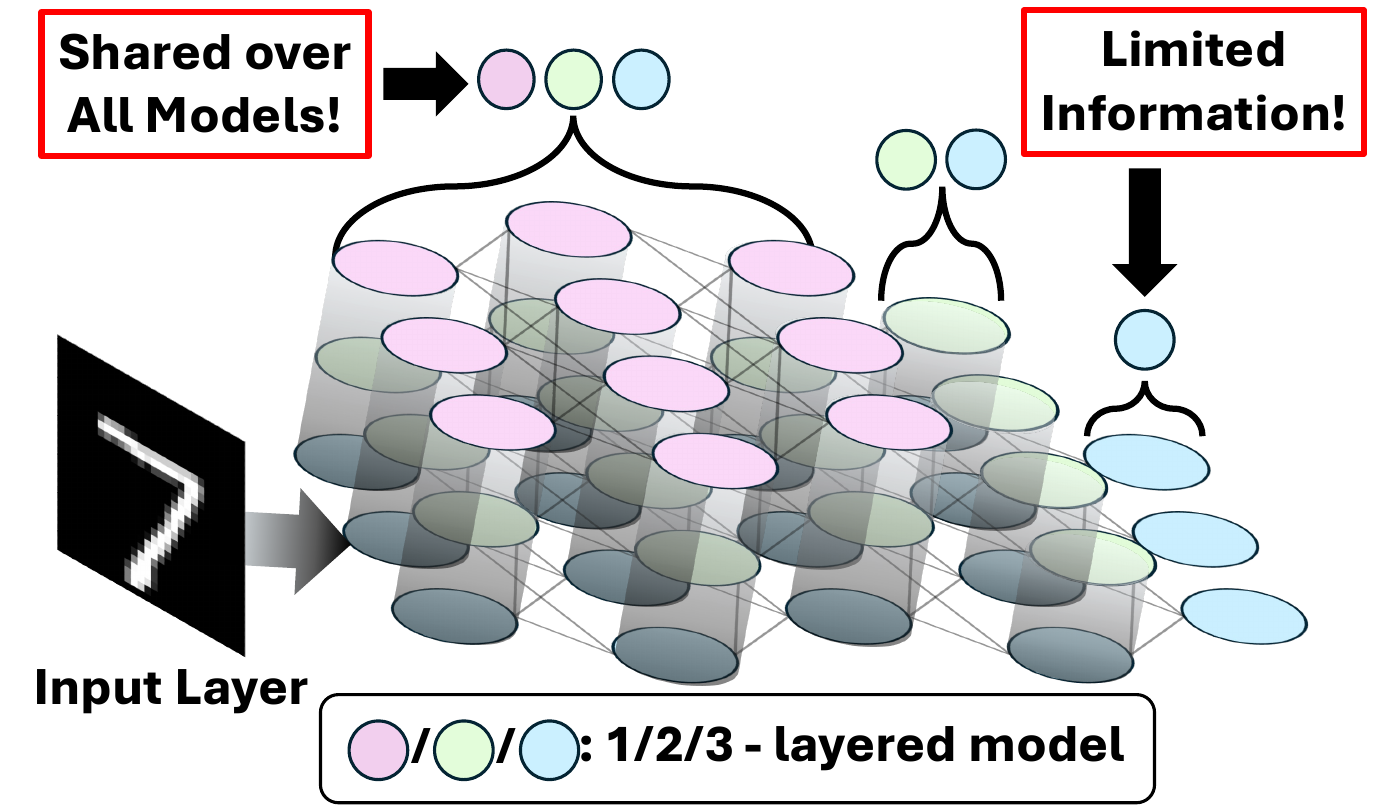}
    \vspace{-2ex}
    \caption{Parameter aggregation in heterogeneous federated learning showing information disparity.}
    \Description{}
    \vspace{-2ex}
    \label{fig:prelim-ME}
\end{figure}

\begin{figure}[!t]
    \centering
    \includegraphics[width=1
    \linewidth]{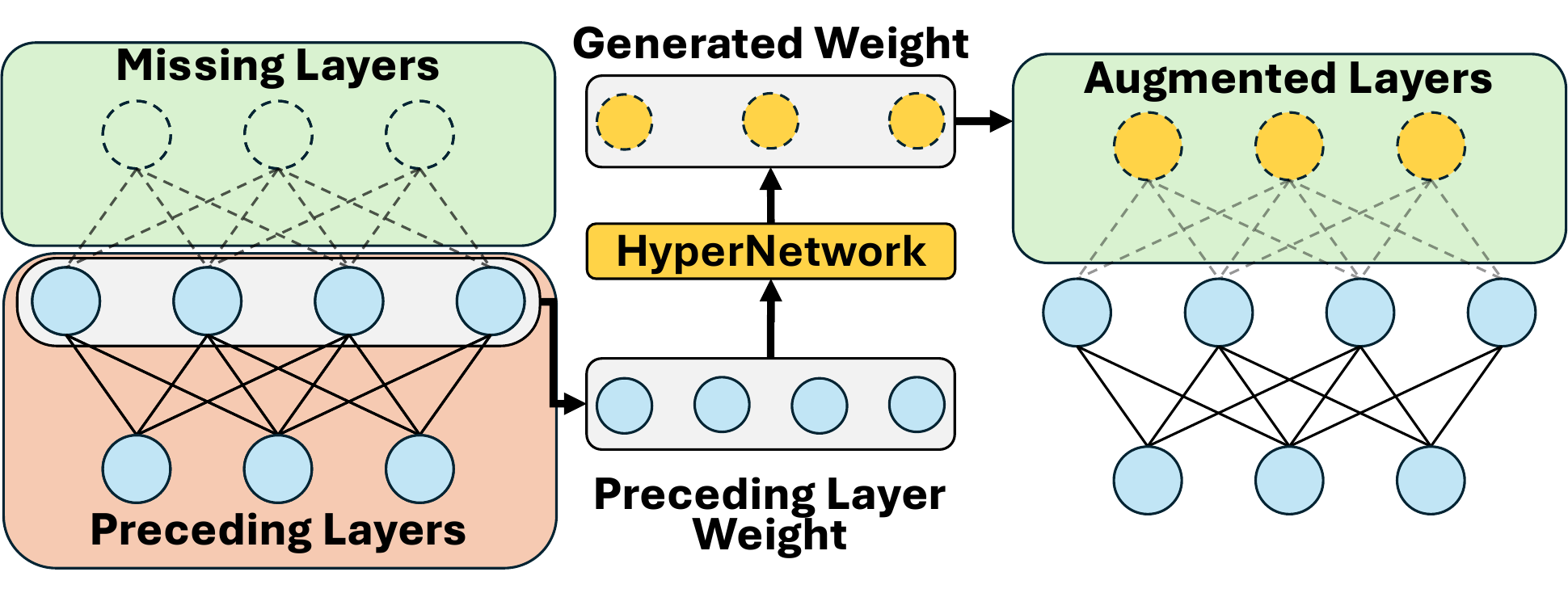}
    \vspace{-2ex}
    \caption{Hypernetwork-based weight generation.}
    \Description{}
    \label{fig:prelim-HN}
    \vspace{-2ex}
\end{figure}

\noindent \textbf{[Issue 2] Per-layer information disparity:}
Figure~\ref{fig:prelim-ME} visualizes the operations of a naive parameter aggregation scheme at the federated learning server for clients operating heterogeneous models. Unlike the earlier layers (closer to the input, depicted in pink) shared across all subnetworks, the number of gathered parameters for the deeper layers can be limited (green and blue). This occurs because all clients execute the earlier (shallow) layers of the model, but only clients with sufficient resources execute the deeper layers. Consequently, when aggregated at the server, the quantities of gathered weights differ for each network layer, depending on how subnetworks are distributed in the network. We define this \textit{per-layer information disparity} as the second issue in realizing depth-wise network splits in federated learning. 
This issue, where specific layers in the neural network receive less information to aggregate, hinders proper model convergence and generalization.

\noindent \textbf{[Solution 2] Hypernetwork-based network parameter generation:} To address per-layer information disparity, we explore leveraging a \textit{hypernetwork for network parameter generation}. Hypernetworks use a set of network parameters to generate the weights for a different network. Figure~\ref{fig:prelim-HN} illustrates how a hypernetwork architecture can be used for federated learning network weight generation. Specifically, the hypernetwork takes the trained parameters from the shallow layers and generates the necessary parameters for deeper layers to facilitate effective aggregation. By doing so, we can generate weights for the $n^{th}$ layer using the $n-1^{th}$ layer weights. This resolves the information disparity issue occurring at the deep layers of a federated learning model. We emphasize that the hypernetwork is trained on the server using model parameters obtained from clients.
We will detail the design and optimization of hypernetworks in Section~\ref{sec:design}.


\vspace{-1ex}
\subsection{Preliminary Study}
\begin{figure*}[!t]
    \centering
     \subfigure[Accuracy performance comparisons for different federated learning configurations and the impact of our potential solutions]{
    \includegraphics[width=0.31\linewidth]{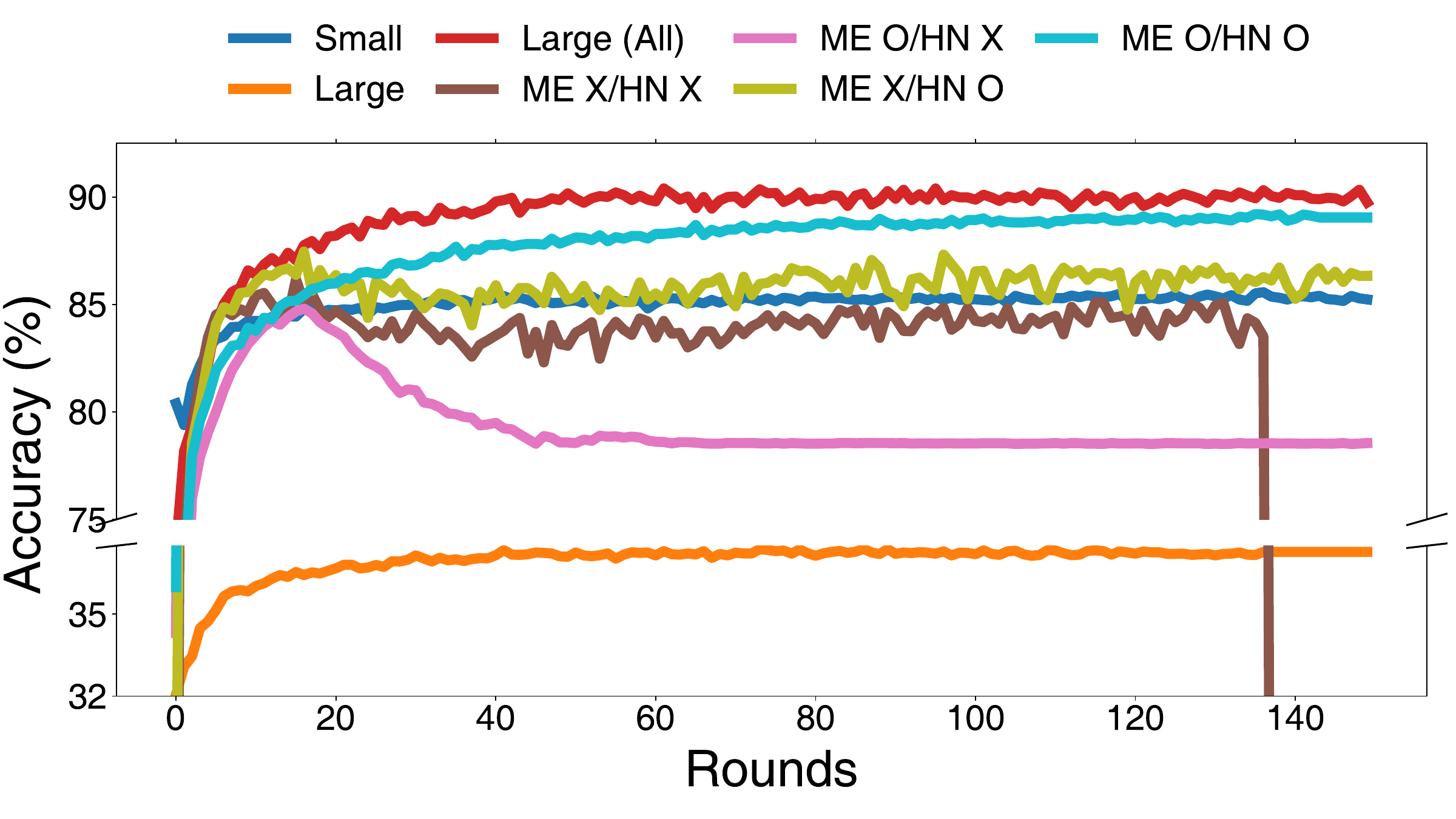}
    \Description{}
    }
    \hspace{1ex}
    \subfigure[CKA values measured at the first layer of a one-layer split (S), two-layer split (M) and full model (L), with (O) and without (X) the multi-exit architecture]{
    \includegraphics[width=0.31\linewidth]{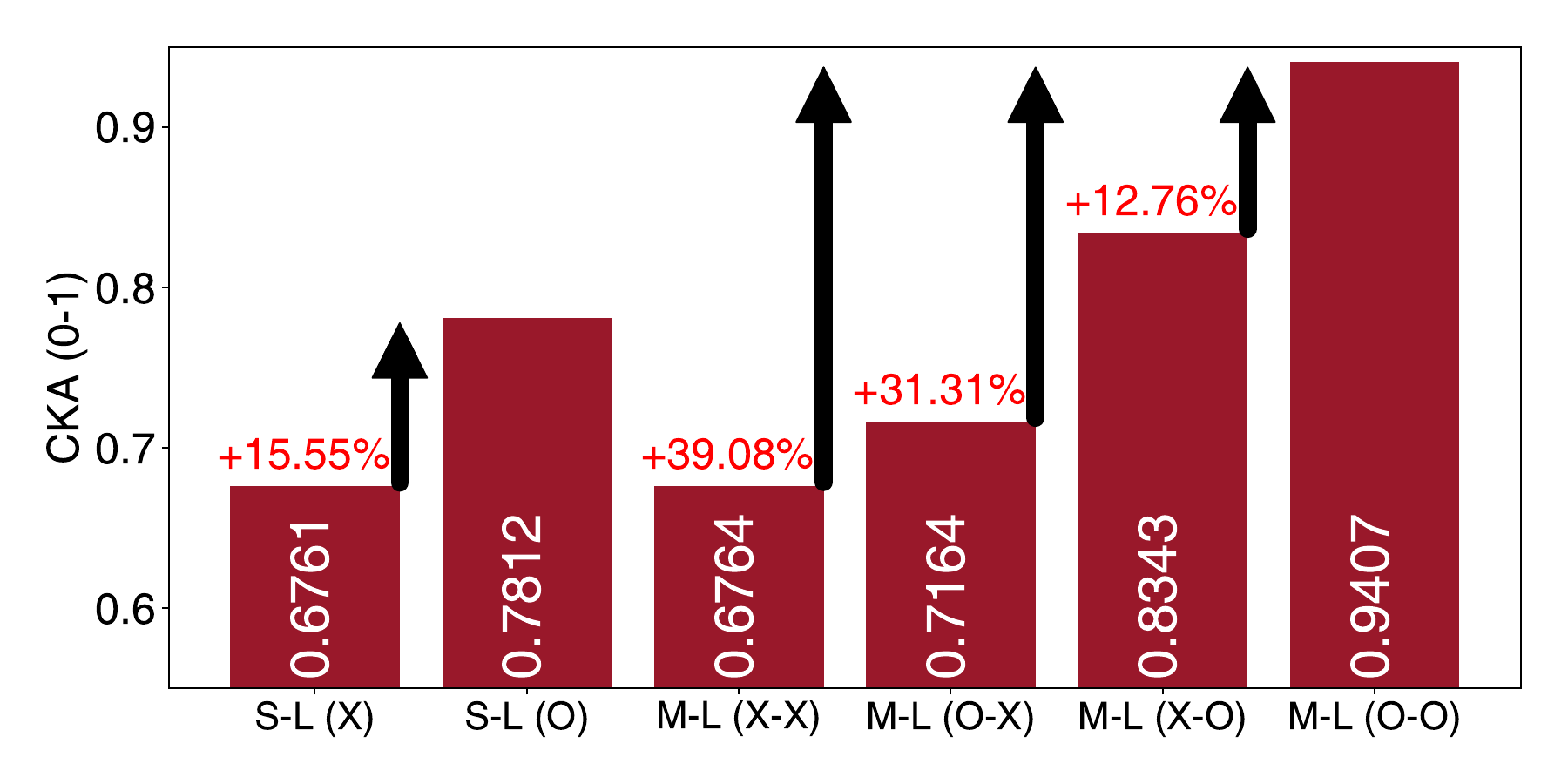}
    \Description{}
    }
    \hspace{1ex}
    \subfigure[Density distribution of trained (ground truth), hypernetwork-generated, and randomly initialized weight values]{
    \includegraphics[width=0.31\linewidth]{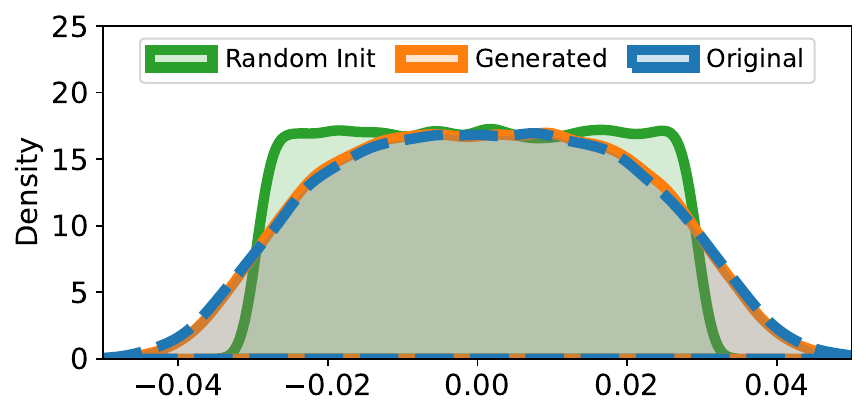}
    \Description{}
    }
    \vspace{-3ex}
    \caption{Accuracy, CKA, and weight value distribution plots for different federated learning configurations and potential solutions proposed in this work.}
    \vspace{-2ex}
    \label{fig:prelim2}
    \Description{}
\end{figure*}

\begin{figure}
    \centering
    \includegraphics[width=0.85\linewidth]{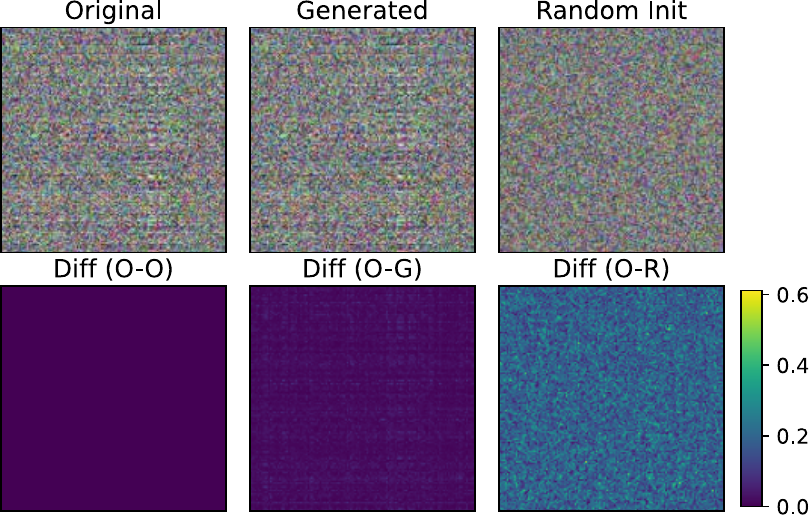}
    \vspace{-1ex}
    \caption{Visualized samples of trained, generated, and randomly initialized parameters.}
    \vspace{-3ex}
    \label{fig:hn-sample}
\end{figure}

We now present a preliminary study designed to examine the potential of exploiting our proposed solutions for heterogeneous federated learning scenarios.

For this study, we utilize the FashionMNIST dataset~\cite{xiao2017fashion}, independently and identically distributed (i.e., IID) across 50 clients, with 10 randomly selected clients participating in each federated learning round. We adopt an IID setup to isolate the impact of our proposed solution from data distribution effects. Our experiments focus on three subnetwork architectures of varying depths derived from a full model consisting of three convolutional layers. Specifically, 17, 17, and 16 clients among the 50 operate 1, 2, and 3-layered models, respectively. We compare against three baselines: (i) all 50 clients using a lightweight single-layer model (``Small'' in Fig.~\ref{fig:prelim2}~(a)), (ii) only capable clients using the full three-layer model (``Large''), and (iii) a non-practical but ideal scenario where all clients use the full three-layer model (``Large (All)'').

Figure~\ref{fig:prelim2}~(a) shows the accuracy results for these configurations. As expected, the ``Large (All)'' scenario, where all 50 clients use the deepest model, achieves the highest performance. However, in the more practical ``Large'' scenario where only a subset (16/50) of clients can use deeper architectures and are the only ones participating, the accuracy remains low and fails to converge, performing worse than the scenario where all clients use the single-layer model (``Small''). These results highlight the importance of optimizing model architectures to accommodate client diversity and maximize client engagement in federated learning processes.

In Figure~\ref{fig:prelim2}~(a) we also plot four additional cases where our proposed solutions in Section~\ref{subsec:3.1} were applied in different combinations. As we allocate clients with their largest-adoptable model, we text cases where neither the multi-exit approach nor the hypernetwork is used (``ME X/HN X''), cases where one of the two is used (``ME O/HN X'' and ``ME X/HN O''), and a final case there both approaches are integrated into the federated learning process (``ME O/HN O''). We use these plots to examine the feasibility of applying our solutions for federated learning scenarios. The results in Figure~\ref{fig:prelim2}~(a) suggest that in the absence of the multi-exit architecture, convergence issues arise due to misaligned feature spaces, while the system faces issues due to per-layer information disparity when without the hypernetwork-based weight generations. \rev{In particular, without a multi-exit architecture, clients with limited resources would provide updates that do not align with those from clients running deeper models; thus, hindering global convergence. Similarly, without hypernetworks, the server cannot effectively gather (enough) weights for deeper layers, leaving gaps in layer-specific knowledge aggregation.} As a result, the case with both multi-exit and hypernetwork (``ME O/HN O'') shows the highest performance, close to that of an ideal case where all clients exploit the full model (``Large (All)''). \rev{These results emphasize the need for integrating both the multi-exit architecture and hypernetwork approaches to ensure robust federated learning across heterogeneous clients.}

Figure~\ref{fig:prelim2}~(b) analyzes feature representation similarity across clients using the Centered Kernel Alignment (CKA) metric~\cite{kornblith2019similarity}. CKA measures similarity between feature spaces of neural networks, where values closer to 1 indicate higher similarity. Effective federated learning models typically show closely aligned feature spaces~\cite{luo2021no}. We compute CKA using the first layer's feature space as a baseline for three configurations: single-layer split (``S''), two-layer split (``M''), and the full model (``L''). Additionally, we examine these configurations with and without multi-exit layers, denoted by ``O'' and ``X'' respectively. For example, ``S-L (X)'' compares the single-layer and three-layer models without multi-exits. 

Without multi-exit layers, the one-layer (S) and two-layer (M) models differ significantly from the full model (L), with CKA values of 0.6761 and 0.6764, respectively. However, incorporating multi-exit architecture improves feature alignment noticeably, yielding CKA values of 0.7812 and 0.9407 for these configurations. \rev{This improved alignment underlines how multi-exit architectures help to preserve feature consistency across clients with varying model depths.} Comparisons like ``M-L (X-O)'' and ``M-L (O-X)'' in Figure~\ref{fig:prelim2}~(b) show considerable differences in feature alignment depending on the presence of multi-exit layers, with CKA values of 0.7165 and 0.8343, respectively. \rev{The improvement in CKA values implies the critical role of multi-exit networks in not only balancing the computational load, but also in maintaining coherent feature spaces.} These findings highlight the effectiveness of multi-exit architectures in depth-wise split networks for federated learning, enhancing feature space alignment and potentially boosting overall model performance.

We evaluate the effectiveness of using a hypernetwork by comparing its generated model weights with the actual weights of a three-layer model (ground truth). The full model was trained on 50 clients, with 40 used for hypernetwork training. We then generated the last layer parameters for the remaining 10 models based on their second-layer parameters. For comparison, we also include randomly generated weights, plotting their distributions in Figure~\ref{fig:prelim2}~(c). The results demonstrate that hypernetwork-based weight generation accurately estimates network weights comparable to the ground truth.
Furthermore, by visualizing the network weights as a matrix and comparing their differences with the ground truth in Figure~\ref{fig:hn-sample}, we can see that the difference in network weight estimation is minimal and that the hypernetwork is capable of intelligently inferring network weights.

Overall, our preliminary results suggest the potential effectiveness of leveraging the multi-exit architecture and hypernetworks for heterogeneous federated learning. \rev{The combination of multi-exit architecture for feature alignment and hypernetworks for filling in missing layers provides a synergistic solution to the challenges posed by client heterogeneity.} Using these results as empirical evidence, we discuss the details of our proposed \system design in the following section.

%% file: sec/4_design.tex
\vspace{-1ex}
\section{System Design}
\label{sec:design}

\subsection{Overview of \system{}}
\label{subsec:overview}
\begin{figure*}[!t]
    \centering
    \includegraphics[width=0.9\linewidth]{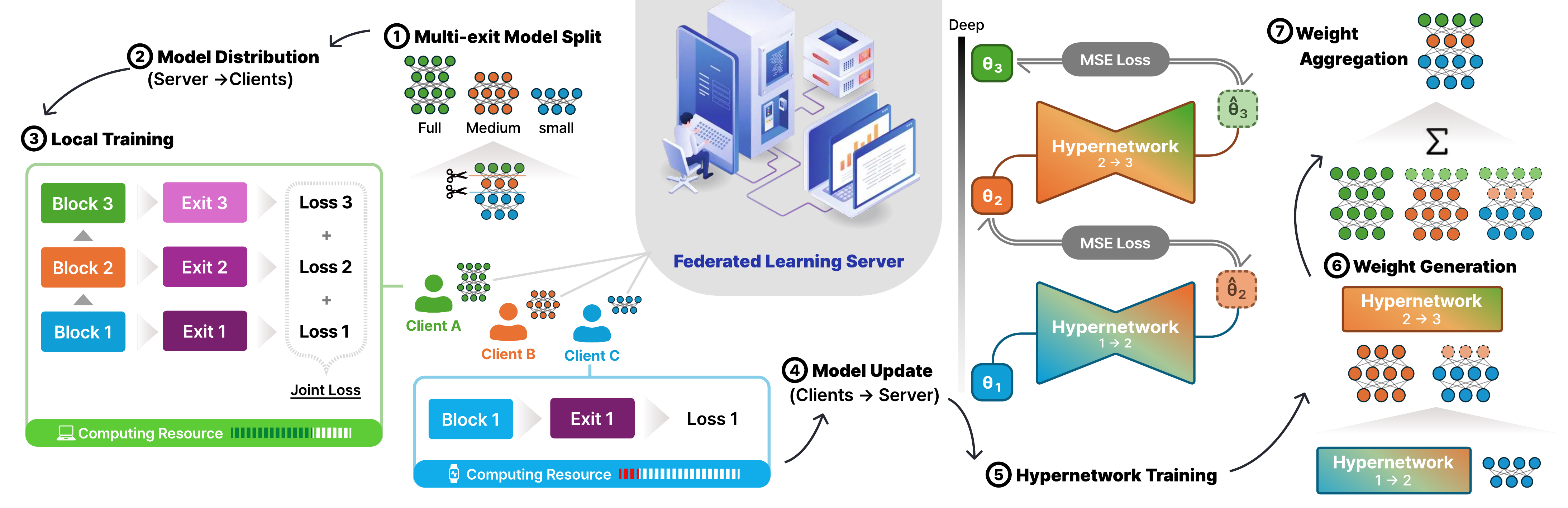}
    \vspace{-2ex}
    \caption{Overview of \system. \system leverages heterogeneous models generated from a multi-exit network architecture with hypernetworks to resolve the per-layer information disparity issue. Best viewed in color.}
    \label{fig:overview}
    \Description{}
\end{figure*}

Our work proposes \system, a federated learning framework for networked systems consisting of clients with heterogeneous computational capabilities. Specifically, \system aims to optimize federated learning across heterogeneous clients and address the challenges detailed in Section~\ref{sec:prelim} by integrating a multi-exit architecture and employing a hypernetwork-based neural network parameter generation scheme. This ensures effective client participation in the federated learning process regardless of the clients' varying computational capabilities. Figure~\ref{fig:overview} provides an overview of \system's design, highlighting its operational process in a single federated learning round:

\begin{enumerate}[leftmargin=*]
    \item The server distributes global model parameters to clients based on their computing resources. \rev{Here, we note that the server can easily acquire information about the clients' computing resources via a simple request.} The \textit{full} global model is configured with multiple exit layers organized in a depth-wise manner (\circled{1}). Clients with higher computational resources receive the full model, while the ones that are limited receive a smaller subset of the model split at earlier exit layers (\circled{2}). (c.f., Sec.~\ref{subsec:multiexitdesign}).

    \item Clients train their models with local data (\circled{3}). \system retains the standard local training procedure unchanged. The key difference lies in clients using heterogeneous models and training them with a joint loss function that incorporates predictions from all available exits. 
    
    \item After local training, each client sends back its trained model parameters to the server (\circled{4}). Clients transmit only the parameters relevant to their assigned model configuration, ensuring efficient parameter aggregation.

    \item The server trains its hypernetworks by leveraging client data that contain sufficient layer information (\circled{5}). The hypernetwork is then used to predict and generate network weights ($\hat{\theta}$ in Fig.~\ref{fig:overview}) for layers suffering from the information disparity issue, particularly the deeper layers or exits of the model. (c.f., Sec~\ref{subsec:hypernetdesign} and Sec.~\ref{sec:hyperimpl}).
    
    \item Finally, the server aggregates the parameters to generate an updated global model based on the client-sent data and the hypernetwork-generated parameters (\circled{6}, \circled{7}). 
    
\end{enumerate}

This structured approach ensures that \system can utilize both powerful and weaker devices, enhancing overall federated learning performance and inclusivity. The following sections detail \system's core components: the multi-exit network architecture and weight generation hypernetworks.

\subsection{Multi-Exit Neural Network Architecture}
\label{subsec:multiexitdesign}

The global model in \system incorporates multiple intermediate classification layers known as ``exit layers,'' distributed in a depth-wise manner within the neural network. Each exit layer, split in a depth-wise manner, functions as a subnetwork of the full model, enabling effective aggregation of client models and supporting heterogeneous federated learning across devices with varying computing capabilities.

The placement and number of exit layers in \system are key design choices. In this work, we use three exit layers positioned at one-third, two-thirds, and the end of the federated learning model to exploit a range of features that the model embeds~\cite{park2023attfl}. This setup allows the categorization of clients into three groups based on their computational capabilities: the first group, with limited resources, uses a subnetwork up to the first exit layer; the second group, with moderate resources, uses up to the second exit; and the third group, with ample resources, uses the full network with all three exits. During local training, clients compute outputs for all exit layers included in their allocated model. For instance, a client with sufficient resources generates predictions from all three exit layers and a resource limited client will compute only up to the first. Each layer's output contributes to the overall loss, summing the losses from all exits—a standard practice in multi-exit neural networks~\cite{laskaridis2021adaptive, teerapittayanon2016branchynet, lee2024recurrent}.
\rev{In this work, we adopt a simple categorization of clients' resource heterogeneity in three levels and accordingly use three exits. However, depending on design choices, the granularity of client resource heterogeneity can adjusted. As we explain later, \system is a general framework that does not depend on the number of exits; thus, the number of exits is not a constraint. Additionally, the optimal placement of exit layers can vary based on the model architecture and the characteristics of the data, and can be determined experimentally by the system designers. In this paper, rather than searching for the optimal exit positions for each dataset, we have assigned the exit positions to uniformly split the model, dividing the model into thirds, to validate a general configuration.}

This multi-exit architecture in \system offers two key advantages. First, it computes losses at both early and final layers, mitigating the vanishing gradient problem and improving training dynamics~\cite{teerapittayanon2016branchynet}. Second, it addresses feature space misalignment from depth-wise model splitting (Section~\ref{sec:prelim}), ensuring that early layers capture both local and global features, promoting better feature alignment and convergence in federated learning.


\system facilitates seamless aggregation of trained model parameters despite heterogeneous network architectures by training and sharing all exit layer parameters. Specifically, \system can easily be integrated with various federated learning schemes like FedAvg~\cite{mcmahan2017communication}, FedProx~\cite{li2020federated}, and SCAFFOLD~\cite{karimireddy2020scaffold} without modifying the local training and aggregation process. \rev{We emphasize that this flexibility further enables \system{} to effectively address the prevalent statistical data heterogeneity observed in real-world usage scenarios. In this work, \system exploits the FedAvg approach, a straightforward method for averaging model parameters during server-side model aggregation.}

\subsection{Efficient Hypernetwork Design for Multi-exit Architectures}
\label{subsec:hypernetdesign}
\begin{figure}[!t]
    \centering
    \includegraphics[width=.95\linewidth]{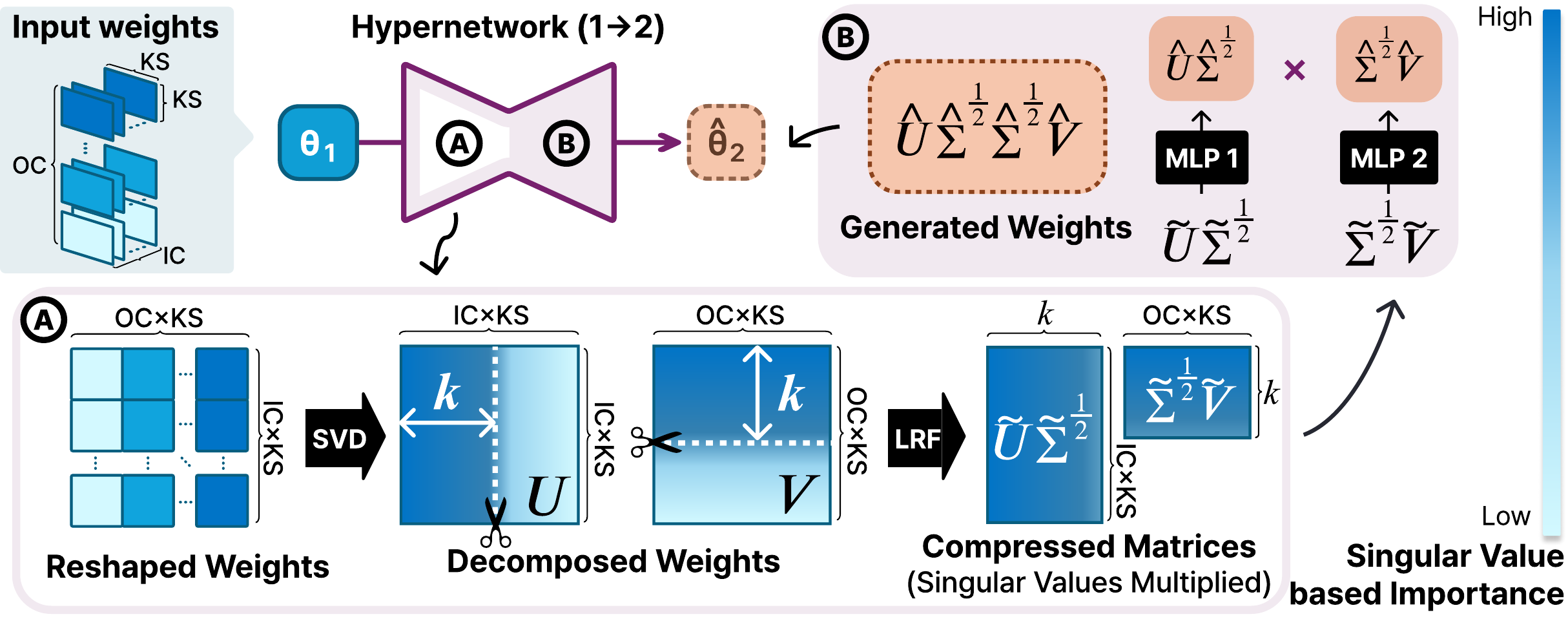}
    \vspace{-2ex}
    \caption{Details on hypernetwork operations.} 
    \label{fig:hypernetworkflow}
    \vspace{-0.5ex}
\end{figure}

Despite its benefits, as discussed in Section~\ref{sec:prelim}, depth-wise model splitting suffers from the information disparity issue at deeper layers. To mitigate this, \system employs a hypernetwork to generate missing weight information. Figure~\ref{fig:hypernetworkflow} illustrates the workflow of this hypernetwork.

The rationale behind using hypernetworks for parameter generation in deeper layers stems from the neural networks' nature as directed acyclic graphs, where computations in later layers depend on results from earlier computations. We hypothesize a nonlinear relationship between parameters of preceding and subsequent layers, which our approach captures through a neural network to predict parameters for deeper exits based on information from earlier exits.

To integrate hypernetworks in federated learning, a straightforward approach involves vectorizing parameters from preceding layers and feeding them into a multi-layer perceptron (MLP) model. The central server, which aggregates parameters from clients with different computational capabilities, possesses information on both preceding and subsequent layer parameters. This allows the server to utilize parameter information from clients computing up to the second and third exits to train the hypernetwork and understand relationships between adjacent exit parameters. However, neural networks typically entail numerous parameters, demanding substantial memory and time for hypernetwork training. \rev{Despite \system's hypernetworks operating on a resource-rich server (a practical configuration for real-world federated learning systems), a na\"ive hypernetwork design can be computationally intensive and memory-consuming, posing practical challenges in real-world deployments, which often makes them impractical with respect to the training time and memory usage (c.f., Section 5.4).} To tackle this, we propose a more efficient hypernetwork architecture that balances effective parameter generation with the computational constraints of federated learning environments.

\system introduces an efficient hypernetwork architecture using a low-rank factorization (LRF) strategy. Instead of vectorizing and reconstructing all model parameters, which is a computationally intensive approach, LRF compresses essential neural network parameters to reduce the computational load and memory requirements. By leveraging the inherent sparsity in neural networks, similar to what is leveraged in pruning techniques, this approach is both feasible and efficient~\cite{he2017channel, frankle2018lottery, evci2022gradient}.

As Figure~\ref{fig:hypernetworkflow} illustrates, \system first reshapes the nearest preceding parameters into a matrix form. For instance, a convolution layer holding parameters with the shape \texttt{IC}$\times$\texttt{OC}$\times\\$\texttt{KS}$\times$\texttt{KS}, where \texttt{IC} denotes the number of input channels, \texttt{OC} denotes the number of output channels, and \texttt{KS} denotes the kernel size, is reshaped into a matrix $\mathbf{A} \in \mathbb{R}^{(\texttt{IC}\times\texttt{KS})\times(\texttt{OC}\times\texttt{KS})}$.

To initiate the compression of this high-dimensional matrix, \system decomposes the matrix-shaped parameters via Singular Value Decomposition (SVD), resulting in three matrices ($\mathbf{A}=\mathbf{U}\mathbf{\Sigma}\mathbf{V}^\top$), namely the left-singular vectors ($\mathbf{U} \in \mathbb{R}^{(\texttt{IC}\times\texttt{KS})\times(\texttt{IC}\times\texttt{KS})}$), right-singular vectors ($\mathbf{V} \in \mathbb{R}^{(\texttt{OC}\times\texttt{KS})\times(\texttt{OC}\times\texttt{KS})}$), and the singular values ($\mathbf{\Sigma} \in \mathbb{R}^{(\texttt{IC}\times\texttt{KS})\times(\texttt{OC}\times\texttt{KS})}$). Note that the singular values $\mathbf{\Sigma}$ indicate the importance of each corresponding $\mathbf{U}$ and $\mathbf{V}$ element. Thus, allowing us to selectively discard less significant parameters, and retaining the most critical components for generating the subsequent layer weights. For example, by discarding the low-rank columns and rows and retaining only the top \( k \) singular values, we obtain reduced singular vectors that consider the scaling factor of the singular values: \(\mathbf{\tilde{U} \tilde{\Sigma}^{\frac{1}{2}}} \in \mathbb{R}^{(\texttt{IC}\times\texttt{KS})\times k}\) and \(\mathbf{\tilde{\Sigma}^{\frac{1}{2}}\tilde{V}} \in \mathbb{R}^{k\times(\texttt{OC}\times\texttt{KS})}\).
The hypernetwork is then trained to predict these compressed singular vectors, approximating the original matrix as \(\mathbf{A} \approx \mathbf{\tilde{U}\tilde{\Sigma}^{\frac{1}{2}}}\mathbf{\tilde{\Sigma}^{\frac{1}{2}}\tilde{V}}\). 

Upon obtaining the two compressed singular value matrices, model parameters are generated efficiently through the straightforward multiplication of these vectors. This method ensures both efficiency and accuracy in parameter generation for deeper layers, addressing the per-layer information disparity problem in federated learning. By leveraging the hypernetwork's ability to capture nonlinear layer relationships, \system enables effective participation from clients with varying computational capacities, promoting robust model convergence and enhancing overall performance.



\subsection{Hypernetwork Implementations}
\label{sec:hyperimpl}


Given the early-exit architecture of \system, which in this paper splits the model into three subnetworks, we employ two hypernetworks: one to predict the second exit parameters from the first exit parameters ($H_{1\rightarrow2}$) and another to predict the third exit parameters from the second ($H_{2\rightarrow3}$). These hypernetworks are trained at the server using parameter samples collected from clients each round. Clients executing the model up to the second exit provide samples for $H_{1\rightarrow2}$, while those executing the full model provide samples for $H_{2\rightarrow3}$. \rev{A briefly mentioned earlier, it is important to note that \system is not limited to a specific number of exits. Our scheme can be generalized to any number of exits, with hypernetworks employed in an auto-regressive manner, where the hypernetwork for each exit ($H_{n\rightarrow n+1}$) predicts the parameters of the next exit based on the current one. This scalable design allows \system to accommodate various levels of model depth and client capabilities, making it adaptable to different deployment scenarios as discussed in Section~\ref{subsec:eval-scalability}.}

Hypernetworks from the previous federated learning round serve as initial values for the next, and new samples are used for further training. We discard samples once they are used to ensure that the hypernetwork generates up-to-date weights while reducing training costs. This approach maintains the effectiveness and efficiency of hypernetworks in adapting to dynamically changing parameters in federated learning.

\system selectively generates only the core parameters of the feature extractor, specifically the convolutional layers, while excluding the batch normalization (BN) and classification layers. This approach is based on two key empirical observations: (i) changes to the BN and classification layers do not significantly enhance global model performance, and (ii) since these layers are closely related to local data, globally generating them can dilute the unique knowledge generated at each client, compromising model personalization. We demonstrate this in Section~\ref{subsec:eval-hypernet}, where we evaluate personalization accuracy after a few epochs of local fine-tuning of the global model with client data.

Once client model parameter data is collected at each federated learning round and the hypernetworks are trained using the operations described above, \system uses the fine-tuned hypernetworks to predict subsequent layer parameters based on the compressed representations. This approach ensures that clients with varying computational capacities can fairly participate in the federated learning process by generating necessary weight information for effective model convergence and improved performance. 

%% file: sec/5_eval.tex
\section{Evaluation}
\label{sec:eval}

\subsection{Experiment Setup}

We perform an extensive set of experiments to validate the performance of \system using three datasets with various comparison baselines, and present details below.


\vspace{0.1in}
\noindent\textbf{Dataset and Model.} \rev{In this work, we use three different datasets for our evaluations and a VGG-based baseline CNN model architecture~\cite{simonyan2014very} with small variations to suit each dataset's characteristics.} We discuss the details of the dataset and model used as follows.

\noindent $\bullet$ \textbf{SVHN dataset~\cite{netzer2011reading}} includes 99,289 labeled \rev{32$\times$32 pixel images} of 10 digits (1-10) extracted from real-world house numbers, offering a diverse representation of physical world images and is well-suited for mobile image classification tasks. The model comprises four convolutional blocks, each with two layers featuring channels (32, 64, 128, 256), followed by max-pooling at the end of each block.

\noindent $\bullet$ \textbf{STL10 dataset~\cite{coates2011analysis}} comprises 10 classes with diverse real-world objects, containing 1.3K samples per class. \rev{Images in STL10 have a relatively higher resolution of 96x96 pixels compared to the SVHN, allowing us to validate \system under different input data conditions.} The model architecture mirrors that used for SVHN, but with channel configurations of 64, 128, 256, and 512.

\noindent $\bullet$ \textbf{UniMiB SHAR dataset~\cite{micucci2017unimib}} 
comprises 4.2K samples from 30 subjects, capturing \rev{three channel (x, y, z-axis) time-series sensor} data from smartphone accelerometers for eight fall types. UniMiB SHAR represents a widely adopted real-world sensing application \rev{exploiting time-series sensor data}. We adopted the same model configuration as the STL10 dataset.

\rev{Note that we choose these datasets to ensure diversity across different real-world scenarios, including image classification tasks with varying levels of complexity, and time-series data for common mobile and embedded sensing applications. The use of such datasets allow us to comprehensively evaluate \system in heterogeneous federated learning environments for various potential use cases. Furthermore, while we present results using model architectures based on CNNs, we note that the proposed approach is applicable to different network architectures (c.f., Section~\ref{sec:discussion}).}



\vspace{0.1in}
\noindent\textbf{Baselines.} For comparison, we utilize five different federated learning schemes: \textbf{(1) FedAvg~\cite{mcmahan2017communication}:} We employ FedAvg as a baseline federated learning approach, which traditionally does not address device heterogeneity. We compare two configurations: in the first, all clients utilize the smallest subnetwork of the multi-exit architecture (FedAvg-S); in the second, all clients utilize the largest model (FedAvg-L). Note that the FedAvg-L configuration is neither realistic nor fair in heterogeneous federated learning, but we test as the upper limit performance; 
%
\textbf{(2) HeteroFL~\cite{diao2020heterofl}}: HeteroFL supports clients with heterogeneous computing resources by leveraging width-wise submodel scaling (compared to our depth-wise splits), where a scaling factor determines the size of the submodel and amount of shared parameters;
%
\textbf{(3) ScaleFL~\cite{ilhan2023scalefl}}: ScaleFL splits the model both width- and depth-wise to reduce the number of model parameters with a preset split ratio, supporting federated learning on devices with heterogeneous computing resources. Additionally, ScaleFL employs self-distillation to optimize each multi-exit;
\textbf{(4) LotteryFL~\cite{li2020lotteryfl}}: Based on the lottery ticket hypothesis~\cite{frankle2018lottery}, LotteryFL achieves personalized model improvements by adaptively pruning model parameters during the client-side local training process to reduce local model capacity;
\textbf{(5) FedRolex~\cite{alam2022fedrolex}}: FedRolex reduces client-side computation by utilizing a subset of the full model, selected using a moving cyclic window. FedRolex supports model variations by adjusting the window size according to client requirements; wider windows enable the use of models with more parameters. 

\vspace{0.1in}


Each of the baselines underwent 300 federated learning rounds with a non-overlapping dataset distributed following a non-IID pattern based on the Dirichlet distribution~\cite{hsu2019measuring}, where $\alpha \in (0, \infty)$ is used to control the degree of data disparity (high $\alpha$ leads to uniformly distributed dataset in terms of the amount of data samples and label distribution). In our experiments, we applied $\alpha=0.5$ as default unless explicitly specified. Local training was performed for five epochs per round with learning rates of 0.005, 0.0005, and 0.0005, and batch sizes of 16, 32, and 128 for the UniMiB, STL10, and SVHN datasets, respectively, using the Adam Optimizer~\cite{kingma2014adam}.


\rev{Throughout the evaluation, by default, \system integrates three exit layers into the federated learning model.} To accommodate heterogeneous clients, we categorize 50 clients into three groups: 17 using a small model with one exit layer, another 17 using a medium-capacity model with two exit layers, and the remaining 16 using the full model with all three exit layers. For UniMiB, we configure 30 clients (10 per group) due to its limited dataset size. \rev{We also test \system{} with varying exit counts (more than three) and locations to further examine the feasibility and generalizability of \system{}.} To assure heterogeneity in the training process, 20\% of the clients were uniformly selected from the three groups to participate in each federated learning round. Except for FedAvg, which does not support device heterogeneity, all other baselines maintained model capacities balanced across client groups. For LotteryFL, we relaxed resource constraints initially to fully train the model for its anticipated performance.


For the hypernetwork, we employed MLP models consisting of two linear layers with ReLU activation for each layer. Training occurred over 25 epochs at the server for each round, utilizing a single-batch approach with a learning rate set to 0.0005 and the Adam Optimizer. \system utilizes SVD for compressing neural network parameters, yielding two singular vectors ($\mathbf{\tilde{U}\tilde{\Sigma}^{\frac{1}{2}}}$ and $\mathbf{\tilde{\Sigma}^{\frac{1}{2}}\tilde{V}}$ in Figure~\ref{fig:hypernetworkflow}), and employs a MLP model for each singular vector. Specifically, we select the top $k$=100 singular values to compress model weights unless otherwise specified.


\subsection{Overall Accuracy Performance}
\label{subsec:eval-overall}
\begin{figure}
    \centering
    \includegraphics[width=.99\linewidth]{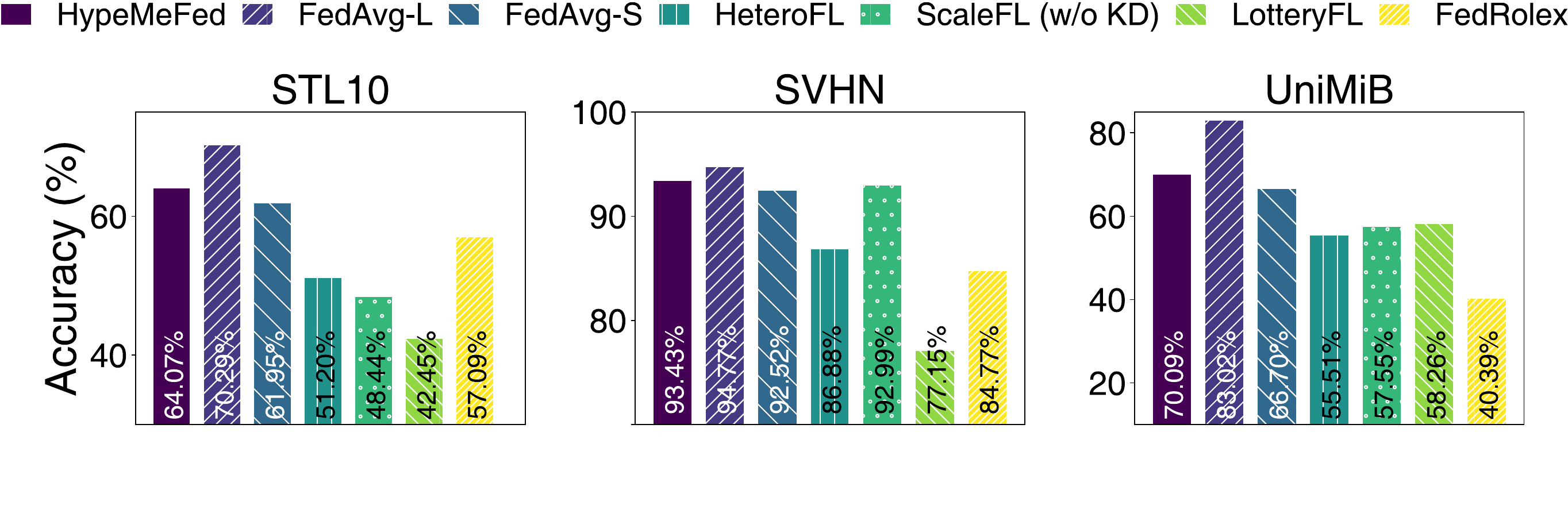}
    \vspace{-4.5ex}
    \caption{Global model accuracy for \system and baselines with varying datasets ($\alpha$=0.5; 300 rounds).}
    \vspace{-2ex}
    \label{fig:acc1}
\end{figure}

Figure~\ref{fig:acc1} plots the global model accuracy following server-side aggregation for \system and baselines. For \system, employing a multi-exit architecture, each model generates multiple predictions per sample, and we evaluate using the final exit prediction. As anticipated, FedAvg-L achieves the highest accuracy across datasets by universally employing the full model, representing an ideal but unrealistic scenario. In contrast, FedAvg-S, sharing a small model to accommodate clients with tight resource limitations, exhibits a consistent accuracy decline. \system outperforms FedAvg-S, effectively narrowing the accuracy gap towards FedAvg-L. 

From Figure~\ref{fig:acc1}, it is evident that baseline methods exhibit inferior performance relative to FedAvg-S, struggling to balance accuracy and support for heterogeneous platforms. Specifically, HeteroFL reduces convolutional layer channels to meet client resource constraints (width-wise scaling), resulting in overly narrow models and poor overall performance. FedRolex, employing a cyclic window approach, requires extensive rounds to cover all model parameters, leading to suboptimal accuracy within a 300-round training period. In its original evaluations~\cite{alam2022fedrolex}, FedRolex required over 1,000 federated rounds for effective convergence, presenting challenges in practical deployments. Additionally, while LotteryFL focuses on improving personalized model accuracy, its global model accuracy falls short. 
%
ScaleFL on one hand, combines depth-wise model splits with width-wise scaling. While the original implementation~\cite{ilhan2023scalefl} also integrates a self-knowledge distillation (KD) operation (c.f., Sec~\ref{sec:relwork}), we noticed that this disturbs the model convergence in some cases (c.f., Sec~\ref{subsec:5.3.1}), showing even lower accuracy; thus, we report results without self-KD. Our findings from ScaleFL suggest that amalgamating hypernetworks to depth-wise model splits offer a more effective approach for supporting heterogeneous federated learning.
We note that this paradigm of heterogenous federated learning schemes performing lower than FedAvg-S is also seen in previous work~\cite{shen2024fedconv}.

\begin{figure}
    \centering
    \includegraphics[width=.99\linewidth]{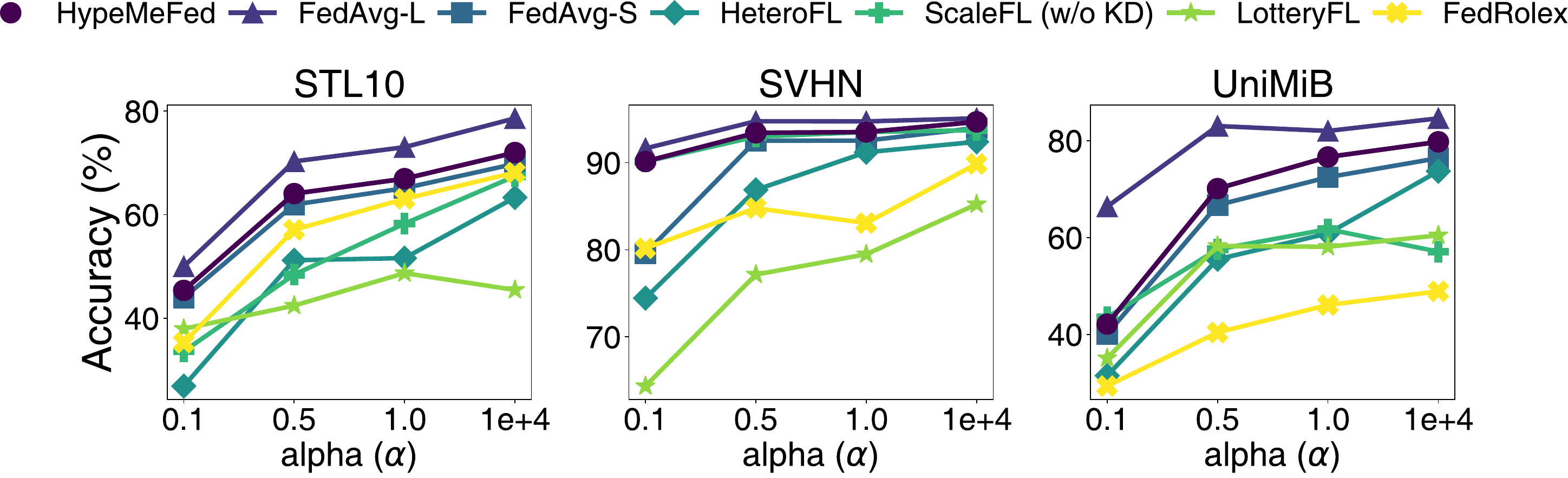}
    \vspace{-2ex}
    \caption{Accuracy with varying non-IID levels ($\alpha$).}
    \label{fig:acc2}
    \vspace{-2ex}
    \Description{}
\end{figure}

Next, while \system targets to address device heterogeneity, we examine its robustness against data heterogeneity by analyzing the accuracy with varying non-IID degrees in the data ($\alpha$). Figure~\ref{fig:acc2} depicts the accuracy for four different $\alpha$ values: 0.1 (strong non-IID), 0.5, 1.0, and 10,000 (IID). The results suggest that as data becomes uniformly distributed, accuracy tends to improve. However, most baselines still fail to outperform FedAvg-S, while \system consistently achieves better or at least competitive model accuracy. \rev{We note that the \system can coexist with advanced aggregation or local training mechanisms to address the issue of device and/or data heterogeneity.}

\system targets to uniformly select participating clients with varying resource capabilities. Nevertheless, while we omit results due to the lack of space, we note that we have empirically confirmed that a non-uniform mix of heterogeneous clients does not affect \system's performance.

\subsection{Impact of Hypernetworks}
\label{subsec:eval-hypernet}
\begin{figure}[t!]
    \centering
    \includegraphics[width=0.99\linewidth]{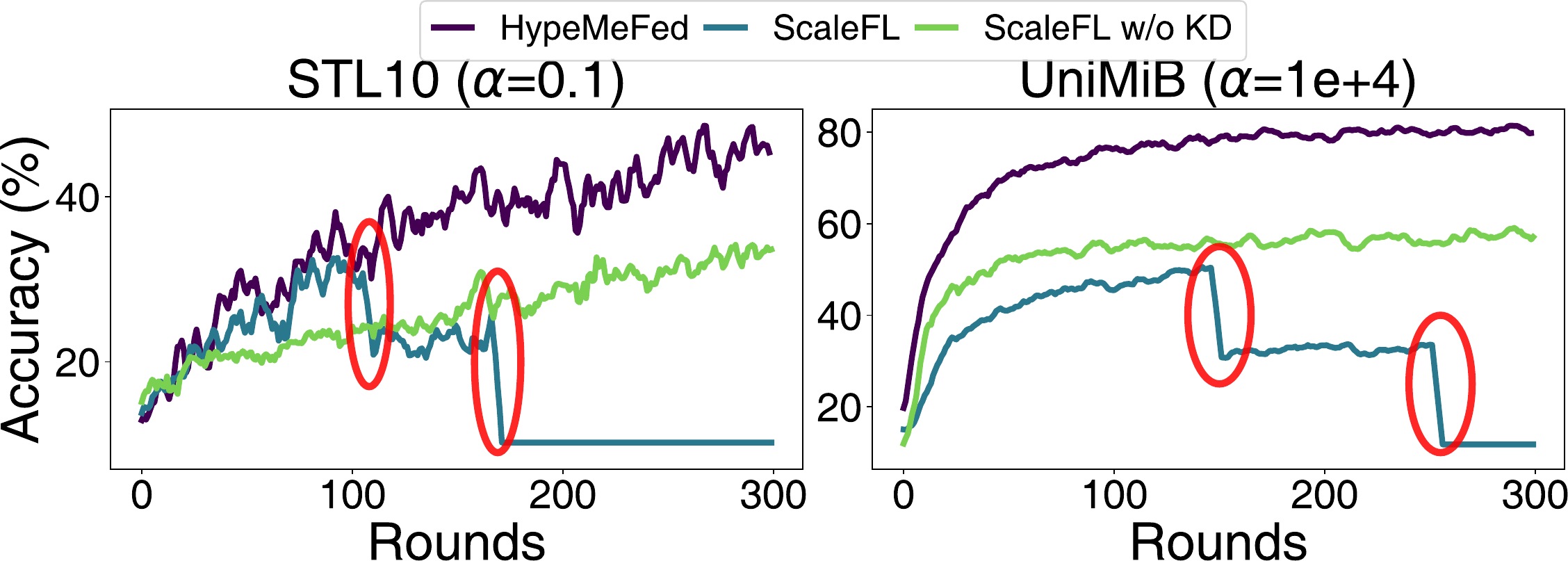}
    \vspace{-2ex}
    \caption{Model accuracy trends with increasing federated learning rounds for \system and ScaleFL.}
    \label{fig:eval-vsscalefl}
    \vspace{-2ex}
    \Description{}
\end{figure}

\subsubsection{Impact of Weight Generation}
\label{subsec:5.3.1}
Given the overall performance improvements that \system brings, we now examine the impact of the hypernetwork on \system's performance. For this, we first examine the global model accuracy trends (at the server) for increasing federated learning rounds for \system, ScaleFL, and ScaleFL without self-knowledge distillation in Figure~\ref{fig:eval-vsscalefl}.
Note that ScaleFL also leverages a multi-exit architecture similar to that of \system, but without explicit weight estimations for the deeper layers. From our evaluations using the STL10 and UniMiB datasets with different $\alpha$, we can notice that ScaleFL fails to properly converge and show accuracy drops during the federated learning operations as the rounds progress (see red highlights in Fig.~\ref{fig:eval-vsscalefl}). By removing the self-KD process, we can see a continuously increasing trend with convergence, but \system eventually converges to a higher accuracy.

\begin{table}[t!]
\begin{adjustbox}{width=0.99\linewidth, center}
\begin{tabular}{|c|ccccc|}
\hline
\multirow{2}{*}{CNN - UniMiB}                                                         & \multicolumn{5}{c|}{Number of Singular Values ($k$)}                                                                                                                                                                                                                                                                                                                                                                                                            \\ \cline{2-6} 
                                                                                      & \multicolumn{1}{c|}{All}                                                                 & \multicolumn{1}{c|}{25}                                                                         & \multicolumn{1}{c|}{50}                                                                & \multicolumn{1}{c|}{100}                                                               & 200                                                               \\ \hline
Accuracy$\uparrow$                                                                    & \multicolumn{1}{c|}{40.44 (0.00\%)}                                                       & \multicolumn{1}{c|}{38.55 (-1.89\%)}                                                              & \multicolumn{1}{c|}{38.76 (-1.68\%)}                                                     & \multicolumn{1}{c|}{38.86 (-1.58\%)}                                                     & \textbf{38.89 (-1.55\%)}                                            \\ \hline
HN size (MB)$\downarrow$                                                              & \multicolumn{1}{c|}{455.34 (0.00\%)}                                                      & \multicolumn{1}{c|}{\textbf{5.60 (-98.77\%)}}                                                     & \multicolumn{1}{c|}{5.68 (-98.75\%)}                                                     & \multicolumn{1}{c|}{5.83 (-98.72\%)}                                                     & 6.14 (-98.65\%)                                                     \\ \hline
\# of parameters (M)$\downarrow$                                                      & \multicolumn{1}{c|}{227.37 (0.00\%)}                                                      & \multicolumn{1}{c|}{\textbf{1.40 (-99.39\%)}}                                                     & \multicolumn{1}{c|}{1.42 (-99.38\%)}                                                     & \multicolumn{1}{c|}{1.46 (-99.36\%)}                                                     & 1.54 (-99.32\%)                                                     \\ \hline
\begin{tabular}[c]{@{}c@{}}Avg training time\\ per epoch(ms)$\downarrow$\end{tabular} & \multicolumn{1}{c|}{\begin{tabular}[c]{@{}c@{}}226.56±0.10\\ (1.00$\times$)\end{tabular}} & \multicolumn{1}{c|}{\textbf{\begin{tabular}[c]{@{}c@{}}91.14±0.54\\ (2.49$\times$)\end{tabular}}} & \multicolumn{1}{c|}{\begin{tabular}[c]{@{}c@{}}91.54±0.76\\ (2.48$\times$)\end{tabular}} & \multicolumn{1}{c|}{\begin{tabular}[c]{@{}c@{}}92.24±0.71\\ (2.46$\times$)\end{tabular}} & \begin{tabular}[c]{@{}c@{}}92.34±0.58\\ (2.48$\times$)\end{tabular} \\ \hline
\end{tabular}
\end{adjustbox}
\caption{Accuracy, hypernetwork size and hypernetwork training time for different $k$.}
\label{tab:eval-hn}
\vspace{-5ex}
\Description{}
\end{table}

\begin{figure}[t!]
    \centering
    \includegraphics[width=.99\linewidth]{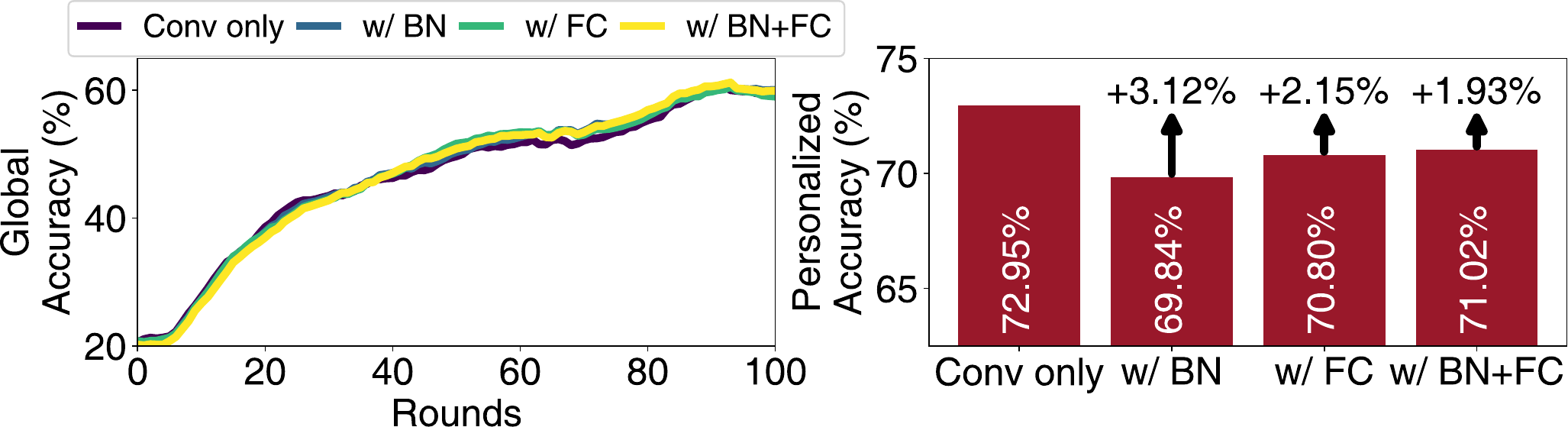}
    \vspace{-2ex}
    \caption{Impact of hypernetwork-based Batch Norm (BN) and Fully Connected (FC) layer weight generation.}
    \label{fig:eval-bnfc}
    \vspace{-2ex}
    \Description{}
\end{figure}

\begin{figure*}[t!]
    \centering
    \subfigure[Weights matrix from original model (ground truth) and weights generated by hypernetwork for a single client with varying $k$]{\includegraphics[width=0.75\linewidth]{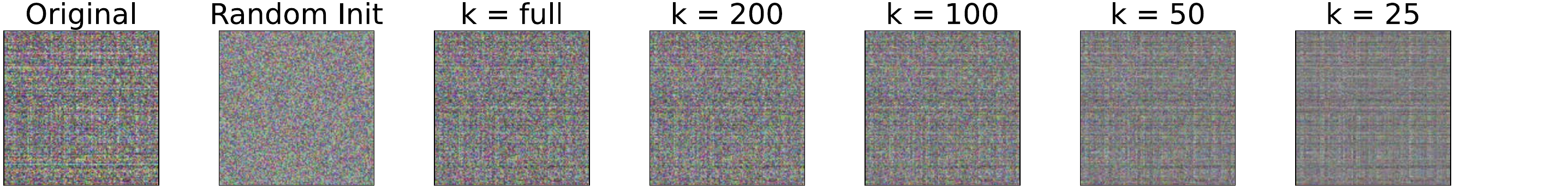}}
    \subfigure[Difference between generated and original weights measured in MAE]{\vspace{-2ex}\includegraphics[width=0.75\linewidth]{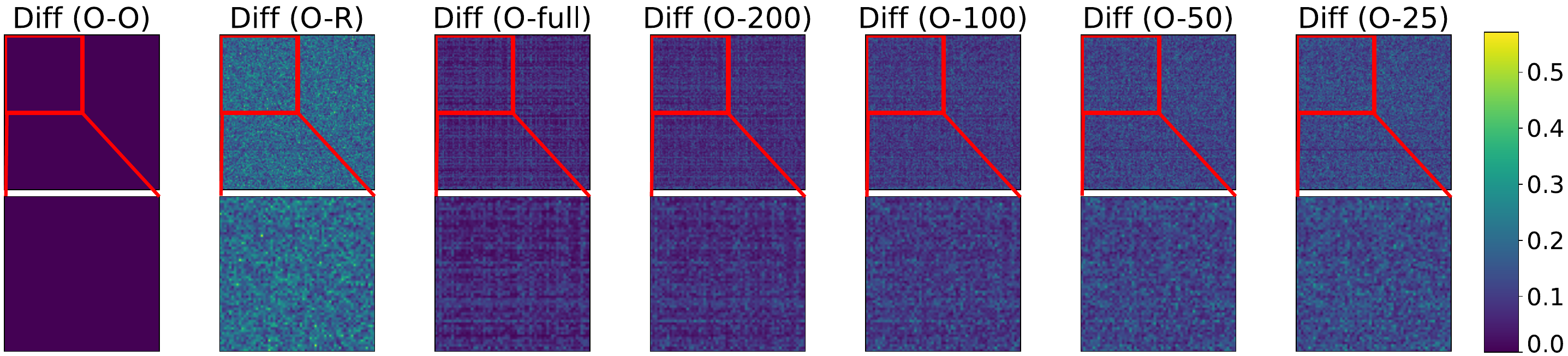}}
    \vspace{-3ex}
    \caption{Visualization plots of weights (and differences) from original model and hypernetwork.}
    \label{fig:eval-ranksample}
    \vspace{-2ex}
    \Description{}
\end{figure*}

\begin{figure}[t!]
    \centering
    \includegraphics[width=0.85\linewidth]{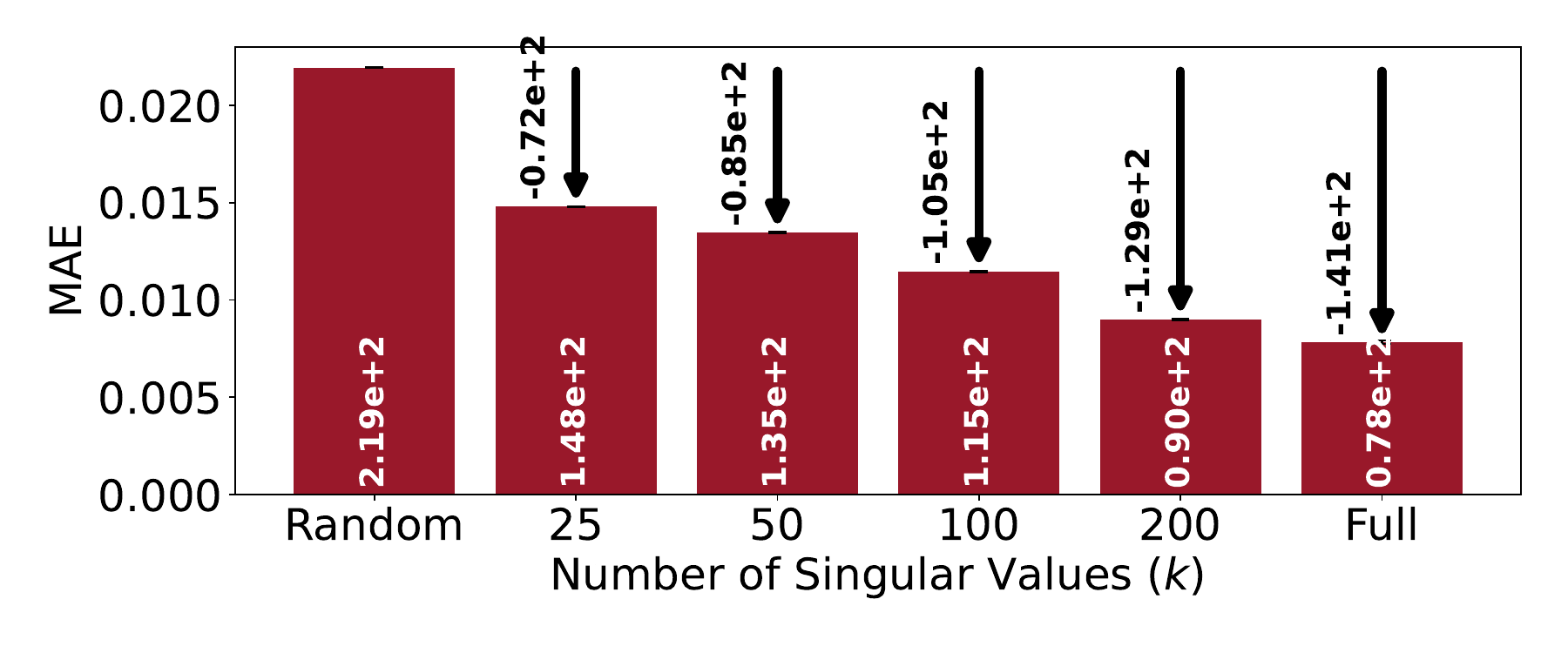}
    \vspace{-2ex}
    \caption{MAE between ground truth and generated weights with varying number of singular values $k$.}
    \label{fig:eval-rankMAE}
    \vspace{-2ex}
    \Description{}
\end{figure}

\subsubsection{Weight Generation for Batch Norm and Fully Connected Layers}

In \system, the hypernetwork focuses on generating weights exclusively for convolutional layers between two exit points. Batch normalization (BN) and fully connected (FC) layers utilize aggregated client information without weight generation to minimize computational and memory overhead. To assess the impact of this design choice on model accuracy, we present results for different hypernetwork-based weight generation scenarios in Figure~\ref{fig:eval-bnfc}.

Overall, as shown in the left of Figure~\ref{fig:eval-bnfc}, with increasing training rounds, the global model accuracy does not exhibit a significant difference between scenarios where the hypernetwork generates BN and FC layer weights and the default scenario where it only generates convolutional layer weights. Additionally, the right plots of Figure~\ref{fig:eval-bnfc} show that personalized accuracy (i.e., accuracy of local model at each client) is highest when weights are generated only for the convolutional layers. 

\rev{We hypothesize that this counter-intuitive observation arises from the increased number of denominators in the aggregation process due to the hypernetwork-generated weights, thereby reducing the impact of model personalization-related information (e.g., locally trained BN and FC layers). In contrast, the feature extractor (i.e., convolutional layers) focuses on extracting universally applicable features across client models~\cite{lubana2022orchestra, xu2021limu, hu2023federated}, maximizing the benefits of weight generation.}


\subsubsection{Impact of Rank on SVD Operations}

In \system, instead of generating weights for the entire model, we employ a low-rank factorization approach based on Singular Value Decomposition (SVD). This method selects the top $k$ ranked parameters for weight generation using the hypernetwork. A larger $k$ enhances model accuracy but increases computational and memory overhead, while a smaller $k$ reduces overhead at the cost of potential performance decrease. Balancing between computational efficiency and model accuracy is crucial in practical federated learning settings with heterogeneous client resources. By tuning $k$, \system can accommodate varying device constraints while maintaining high performance and efficient resource utilization.


To examine the impact of $k$ we take a deeper look into this tradeoff using Table~\ref{tab:eval-hn}. As the results in the table show for the CNN baseline model and the UbiMiB dataset, the model compression process does show a performance degradation in accuracy performance. Nevertheless, we can notice that this loss is minimal even for a very small $k$ (e.g., 1.89\% loss for $k$=25). At the same time, the reduction in parameters and the size of the hypernetwork reduces by nearly two orders of magnitude. Naturally, this reduces the computation latency to a minimal level as well. 

On a different dimension, Figure~\ref{fig:eval-ranksample} visualizes the convolutional layer parameters of the CNN model trained with the UniMiB dataset from a client (i.e., ``original'') and compares them with random weights and hypernetwork-generated weights with different $k$ values. Figure~\ref{fig:eval-ranksample}~(a) shows the actual weight values, while Figure~\ref{fig:eval-ranksample}~(b) displays the differences against the original weights. As illustrated, a hypernetwork with a higher $k$ exhibits fewer differences from the original weights, whereas even a small $k$ demonstrates noticeable improvements over random weights.


To validate this observation, Figure~\ref{fig:eval-rankMAE} shows the Mean Absolute Error (MAE) of the weights for different $k$. Quantitatively, the results here confirm that as $k$ increases, the hypernetwork-generated weights closely approximate their respective original values, with the scenario where the hypernetwork generates all weights showing the lowest MAE. This indicates that leveraging a hypernetwork with an appropriate $k$ effectively balances computational efficiency and model accuracy, thereby supporting robust federated learning across heterogeneous clients.

\subsection{Applicability to Large Models}
\label{subsec:eval-scalability}
\begin{figure}[t!]
    \centering
    \includegraphics[width=0.85\linewidth]{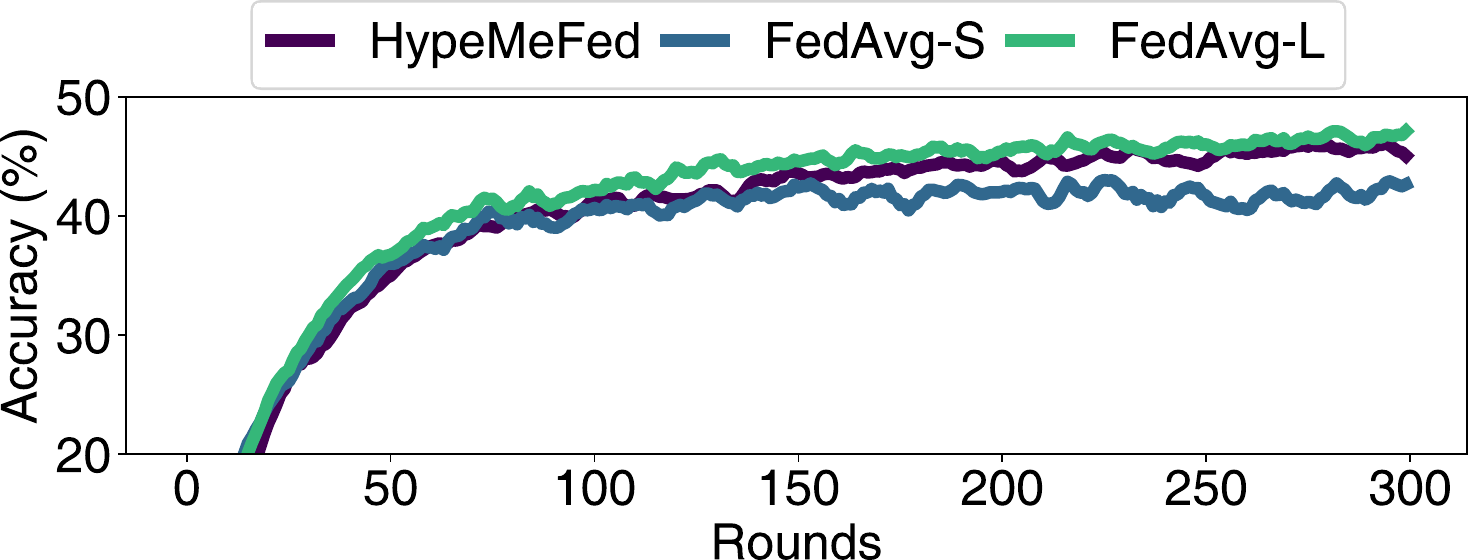}
    \vspace{-2ex}
    \caption{Performance of \system with ResNet18.}
    \label{fig:eval-resnet}
    \vspace{-3.5ex}
    \Description{}
\end{figure}

Experiments conducted so far utilize a relatively small-sized model with a limited number of convolutional layers. The next question we answer is whether \system can scale to larger and deeper models. \rev{Additionally, we also conduct experiments with varying \textit{the numbers of exits} to explore whether it remains effective when models are split in different granularities.} For this, we examine the performance with (i) a more complex model with extensive parameters and (ii) models with deeper layers. Specifically, to test a model with extensive parameters, we conducted experiments with ResNet18~\cite{he2016deep}—a relatively large model with 11.5M parameters, featuring a residual connection architecture. We used the CIFAR100 dataset~\cite{krizhevsky2009learning}, containing 50,000 image samples across 100 categories. Three intermediate multi-exit layers were added to the baseline ResNet18 model, splitting it depth-wise into three segments.

Figure~\ref{fig:eval-resnet} plots the global model accuracy trend with increasing federated learning rounds. The results suggest that \system is applicable to larger models, as it successfully converges and shows improved model accuracy compared to the baseline where all users utilize the smallest model (FedAvg-S in Fig.~\ref{fig:eval-resnet}). In fact, \system achieves performance comparable to the upper bound assuming all clients can utilize the largest model (i.e., FedAvg-L). Here, training the full-rank hypernetwork for ResNet18 requires more than 44.74~GB ($\sim$5.59B parameters) of memory, which is infeasible and computationally extensive. However, with $k$=100 for the hypernetwork optimizations, only 113.28~MB ($\sim$14M parameters) was needed, reducing the memory usage by 99.87\%. \rev{These results demonstrate that \system can effectively scale to larger models while maintaining high accuracy and significantly reducing memory overhead, showcasing its applicability in real-world deployments.}

Note that, \system adopts an auto-regressive approach to generate subsequent layer weights, utilizing either preceding layer weights or previously generated weights. Thus, investigating the influence of model depth, particularly the number of exit layers, becomes pivotal. To explore this, we employed a basic CNN model with 10 exit layers on the FashionMNIST dataset, accommodating varying levels of client device capabilities. We use this dataset given the number of classes and subnetworks that need to be generated for deep networks. Training involved five clients per model capacity (i.e., number of exits in the model), encompassing a total of 50 clients participating in the training phase.

%

\begin{figure}[t!]
    \centering
    \subfigure[Normalized MAE between generated weights and original weights in different layers.]{\includegraphics[width=0.85\linewidth]{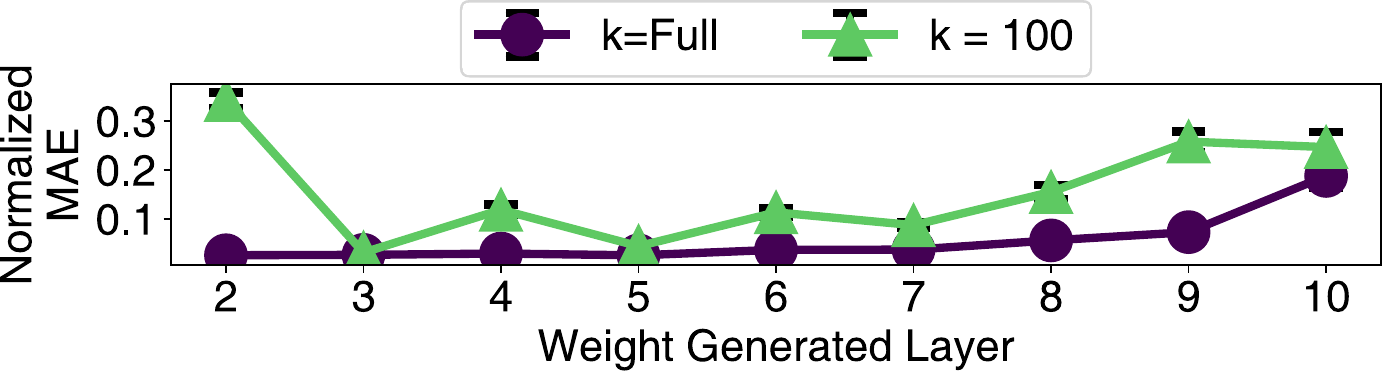}}
    \hspace{1ex}
    \subfigure[MAE between generated weights and original weights in the final layer with respect to the initial layer.]{\includegraphics[width=0.78\linewidth]{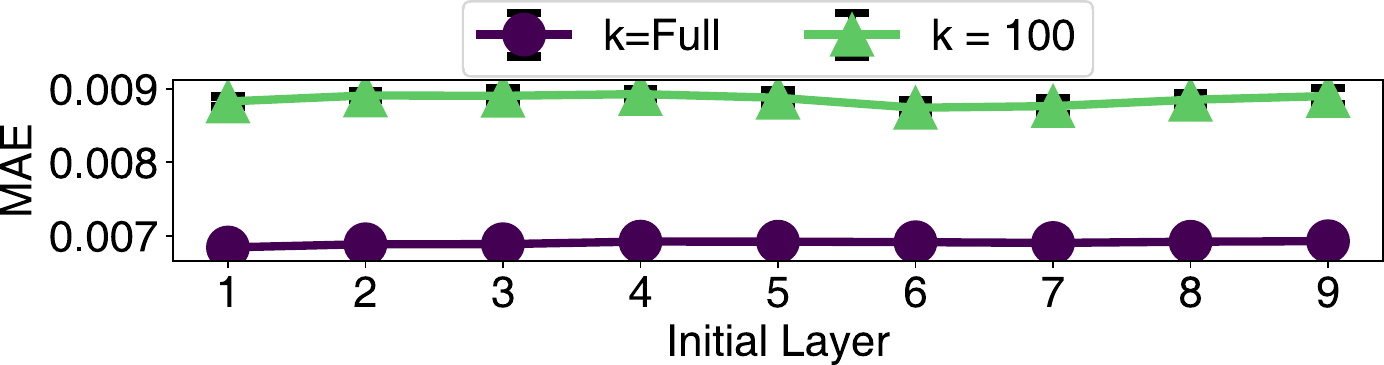}}
    \vspace{-2ex}
    \caption{Weight generation errors in deep model.}
    \vspace{-3ex}
    \label{fig:eval-depth}
    \Description{}
\end{figure}

Figure~\ref{fig:eval-depth}~(a) plots the normalized Mean Absolute Error (MAE) between the original and generated weights with respect to layer depth. Note that we present normalized MAE given that the magnitude of the weight norm differs across layers. 

As the results show, the normalized MAE tends to increase with layer depth despite generating all weights ($k$=Full). This trend likely arises from the challenge of accurately capturing global features in deeper layers, as discussed in previous studies~\cite{luo2021no, lee2024flex}. Despite this observation, \system achieves a higher global accuracy of 89.47\% (+3.71\%) compared to the FedAvg-S (85.76\%), where all clients use the lightest (1-exit) model. Figure~\ref{fig:eval-depth}~(b) plots the MAE between original and generated weights at the final convolution layer. Given an auto-regressive generation approach, where we predict \(n^{th}\) layer weights using \((n-1)^{th}\) layer weights, we assess the accumulated error by comparing MAE across different initial layers used to generate the final layer. Notice that the MAE remains consistent across initiating layers, indicating the robustness of the proposed hypernetwork. \rev{These results confirm that \system's hypernetwork-based approach effectively handles deeper models and larger networks with more exit points, maintaining both accuracy and computational efficiency.}

\subsection{Evaluation on Real-World Testbed}
\label{subsec:eval-platform}
\begin{figure}[t!]
    \centering
    \includegraphics[width=1.0\linewidth]{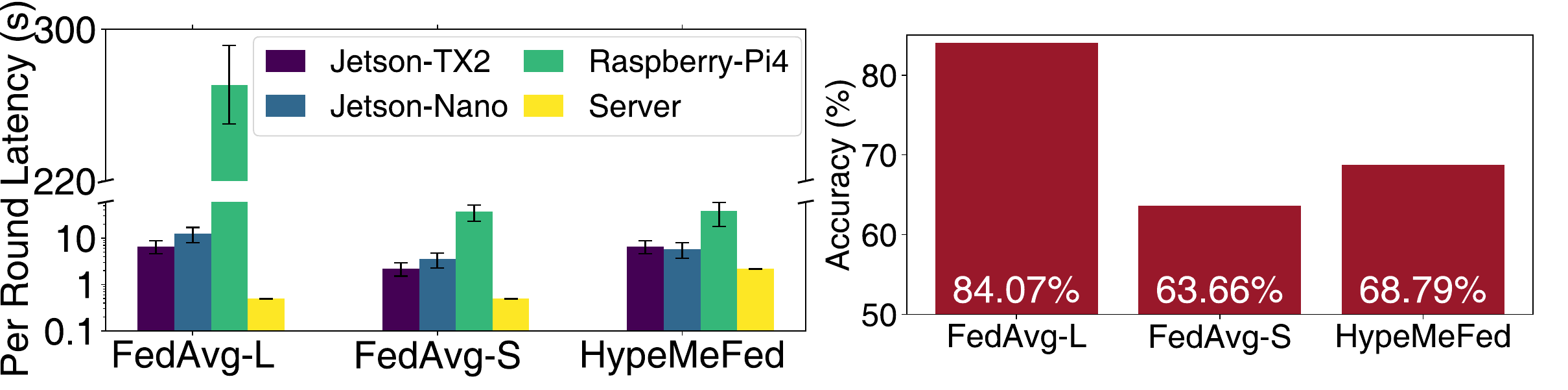}
    \vspace{-2ex}
    \caption{Per round latency and global model accuracy measured from heterogeneous client testbed.}
    \vspace{-3.5ex}
    \label{fig:eval-testbed_performance}
    \Description{}
\end{figure}

\begin{figure}[t!]
    \centering
    \includegraphics[width=0.85\linewidth]{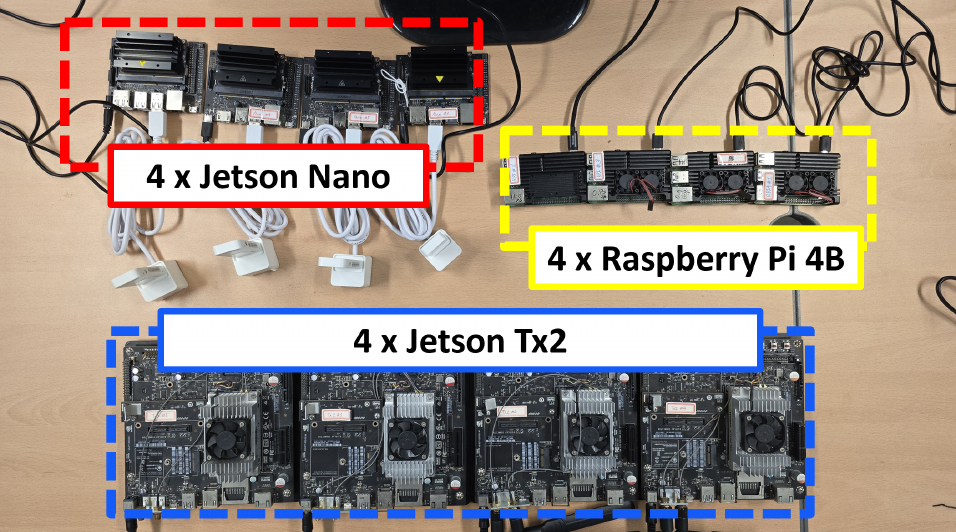}
    \vspace{-2ex}
    \caption{\rev{Picture of our heterogeneous federated learning testbed environment.}}
    \vspace{-3.5ex}
    \label{fig:eval-testbed_setup}
    \Description{}
\end{figure}

To demonstrate the practicality of \system in real-world scenarios, we evaluated \system on a testbed of heterogeneous embedded platforms: 4 Raspberry Pi 4, 4 Nvidia Jetson Nano, and 4 NVIDIA Jetson TX2, totaling 12 devices with varying computational resources—weak, moderate, and ample, respectively \rev{(c.f. Figure~\ref{fig:eval-testbed_setup}).} Additionally, we used a server with an Nvidia RTX 3090 GPU, an Intel i9-K@3.6GHz CPU, and 64GB RAM. We test the UniMiB dataset with a VGG model, where each client holds a non-overlapping dataset distributed following the Dirichlet distribution ($\alpha$=0.5). FedAvg-S and FedAvg-L universally used single- and three-layered models, respectively, while clients in \system used models suitable for each device's resources. Unless otherwise specified, all configurations matched those mentioned in Section~\ref{sec:eval}.

Figure~\ref{fig:eval-testbed_performance} plots the latency per federated learning round for different client platforms along with the server-side latency (left) and the global model accuracy for FedAvg-L, FedAvg-S, and \system (right). \rev{As expected, FedAvg-L achieves high accuracy, but the Raspberry Pi shows prolonged latency in processing this model, potentially delaying the overall federated learning process.} In FedAvg-S, latency is kept low, but using a single-layer model causes a 20.41\% accuracy drop. By applying \system and adaptively configuring models for each device, we balance training latency and accuracy, outperforming FedAvg-S by 5.13\%. Note that a naive hypernetwork induces 3.9~sec of server computation time and 5.6~GB of memory overhead, while our optimizations ($k$=100) reduce this to 2.1~sec (1.86$\times$ speedup) and 100.8~MB (98.22\% reduction), agreeing with our previous results.
%
%
Overall, our testbed validations confirm that \system can be a practical and efficient framework for supporting federated learning on heterogeneous clients.

%% file: sec/6_discussion.tex
\section{Discussions}
\label{sec:discussion}

We now compile a set of notable discussion points for future research that we have identified through our research. 

\begin{itemize}[leftmargin=*]

    \item \textbf{Optimizing hypernetwork architectures and performance.} In this work, we demonstrated the feasibility of exploiting hypernetworks in federated learning through \system. While \system effectively addresses the complexity issues by leveraging SVD-based low-rank factorization, research on hypernetworks is still in its early stages. We identify two potential research directions. First, \system utilizes the same hypernetwork architecture across all layers. Exploring optimal configurations for each layer may further enhance hypernetwork performance. Second, despite reduced complexity, training hypernetworks introduces server-side latency, delaying the federated learning process. Developing efficient hypernetwork training algorithms or suitable alternatives could mitigate this issue and improve overall system performance. \rev{In addition, to better leverage hypernetworks in device-heterogeneous federated learning and understand their performance, better-defined metrics for evaluating hypernetwork performance may be needed. For example, beyond traditional metrics such as accuracy and MAE (as used in this work), it would be helpful to assess the contextual meaningfulness of the generated weights. Furthermore, evaluation metrics that help identify the sweet spot within the hypernetwork complexity and required resources trade-off could ensure a more seamless integration of hypernetworks into federated learning.}


    %
    \item \textbf{Diversifying model architectures.} In this work, we focused our evaluations on CNN-based architectures. However, \rev{we emphasize that \system is inherently model-agnostic, and its design can be extended to support a wide range of neural network architectures beyond CNNs. Specifically, our hypernetwork-based weight generation approach can be applied to other architectures, such as MLPs, RNNs, and LSTMs, by leveraging the layer-wise relationships within the model to generate missing or incomplete weights. This flexibility allows \system to be applicable to various federated learning scenarios involving diverse model types: enabling it to be used in broader applications ranging from natural language processing tasks with RNNs or LSTMs to structured data processing using MLPs.} As we expand our research to include more complex neural network architectures, such as Transformers, we anticipate new challenges. \rev{For instance, Transformer-based architectures, known for their attention mechanisms and high parameter complexity, may increase the demands on hypernetwork capacity and efficiency. The dynamic and dense inter-dependencies of such models may require more sophisticated hypernetwork designs capable of capturing both local and global relationships across layers, while still minimizing memory and computational overhead. Addressing these challenges could open up new possibilities for deploying \system in cutting-edge applications such as NLP, vision-language cooperation applications, and other domains that require sophisticated model architectures. \system’s model-agnostic nature holds significant potential for future federated learning environments, allowing it to adapt to evolving model architectures and application domains.}

    \item \textbf{\system as an integrated system.} \system can be used as an orthogonal scheme with other federated learning frameworks. Specifically, \system does not employ pruning or quantization to the model, allowing for straightforward integration with various schemes. However, this work primarily focuses on exploring the feasibility of \system, and we did not investigate the potential benefits of such integration as a comprehensive solution. Moving forward, we believe that enhancing \system could involve leveraging our proposed techniques to address both systemic and statistical heterogeneity effectively.
\end{itemize}


%% file: sec/7_conclusion.tex
\vspace{-1ex}
\section{Conclusion}
\label{sec:conclusion}

This work presents \system, a novel framework designed to address the challenges of device heterogeneity in federated learning. By combining multi-exit network architectures with hypernetwork-based model weight generation, \system effectively resolves the feature space misalignment between heterogeneous models and the per-layer information disparity issue during weight aggregation. Furthermore, the proposed low rank factorization-based hypernetwork optimization minimizes computational and memory overhead, making hypernetworks feasible (and applicable) for real-world use. Our evaluations with GPU emulations and a real-world heterogeneous embedded platform testbed confirm that \system significantly improves model accuracy and can effectively engage heterogeneous clients in federated learning, proving its practicality in real applications. 